\documentclass[runningheads]{llncs}
\usepackage[T1]{fontenc}

\usepackage{graphicx}
\usepackage{macros}

\usepackage{subfig}
\usepackage{xspace}
\usepackage[
   open,
   openlevel=5,
   atend
 ]{bookmark}
\graphicspath{{figures/}}

\usepackage[normalem]{ulem}

\usepackage{color,soul}

\begin{document}
\title{Offline Task Assistance Planning on a Graph: \\ Theoretic and Algorithmic Foundations}
\titlerunning{Offline g-TAP: Theoretic and Algorithmic Foundations }
%
\author{Eitan Bloch\inst{1} \and
Oren Salzman\inst{1}}
\authorrunning{E. Bloch and O. Salzman}
%
\institute{Technion — Israel Institute of Technology Haifa, Israel.
\email{\{eitanbloch,osalzman\}@cs.technion.ac.il}}

\maketitle              
\begin{abstract}
In this work we introduce the problem of task assistance planning where we are given two robots~$\taskR$ and~$\assistR$.
The first robot,~$\taskR$, is in charge of performing a given task by executing a precomputed path.
The second robot,~$\assistR$, is in charge of assisting the task performed by~$\taskR$ using on-board sensors. 
The ability of~$\assistR$ to provide assistance to~$\taskR$ depends on the locations of both robots. Since~$\taskR$ is moving along its path,~$\assistR$ may also need to move to provide as much assistance as possible.
The problem we study is how to compute a path for~$\assistR$ so as to maximize the portion of~$\taskR$'s path for which assistance is provided.
We limit the problem to the setting where~$\assistR$ moves on a roadmap which is a graph embedded in its configuration space and show that this problem is \NP-hard.
Fortunately, we show that when $\assistR$ moves on a given path, and all we have to do is compute the times at which $\assistR$ should move from one configuration to the following one,  we can solve the problem optimally in polynomial time.
Together with carefully-crafted upper bounds, this polynomial-time algorithm is integrated into a Branch and Bound-based algorithm that can compute optimal solutions to the problem outperforming baselines by several orders of magnitude.
We demonstrate our work empirically in simulated scenarios containing both planar manipulators and UR robots as well as  in the lab on  real robots.

\keywords{Motion and Path Planning \and Algorithmic Completeness and Complexity.}
\end{abstract}

\section{Introduction}
\label{sec:into}
In this work we introduce the problem of \myemph{task assistance planning} (TAP) where we are given two robots~$\taskR$ and~$\assistR$.
The first robot,~$\taskR$, which we call the \myemph{task robot}, is in charge of performing a given task by executing a precomputed path.
The second robot,~$\assistR$, which we call the \myemph{assistance robot}, is in charge of assisting the task performed by~$\taskR$ using on-board sensors.\footnote{
Importantly, by assistance we mean assistance that does not affect the task itself such as communication, micro-control and visual monitoring.
} 
The ability of~$\assistR$ to provide assistance to~$\taskR$ depends on the locations of both robots. Since~$\taskR$ is moving along its path,~$\assistR$ may also need to move to provide as much assistance as possible. In its simplest form, the problem calls for computing a path for~$\assistR$ so as to maximize the portion of~$\taskR$'s  path for which assistance is provided.

\begin{figure}[t!]
    \centering
    \subfloat[]{\includegraphics[height=2.23cm]{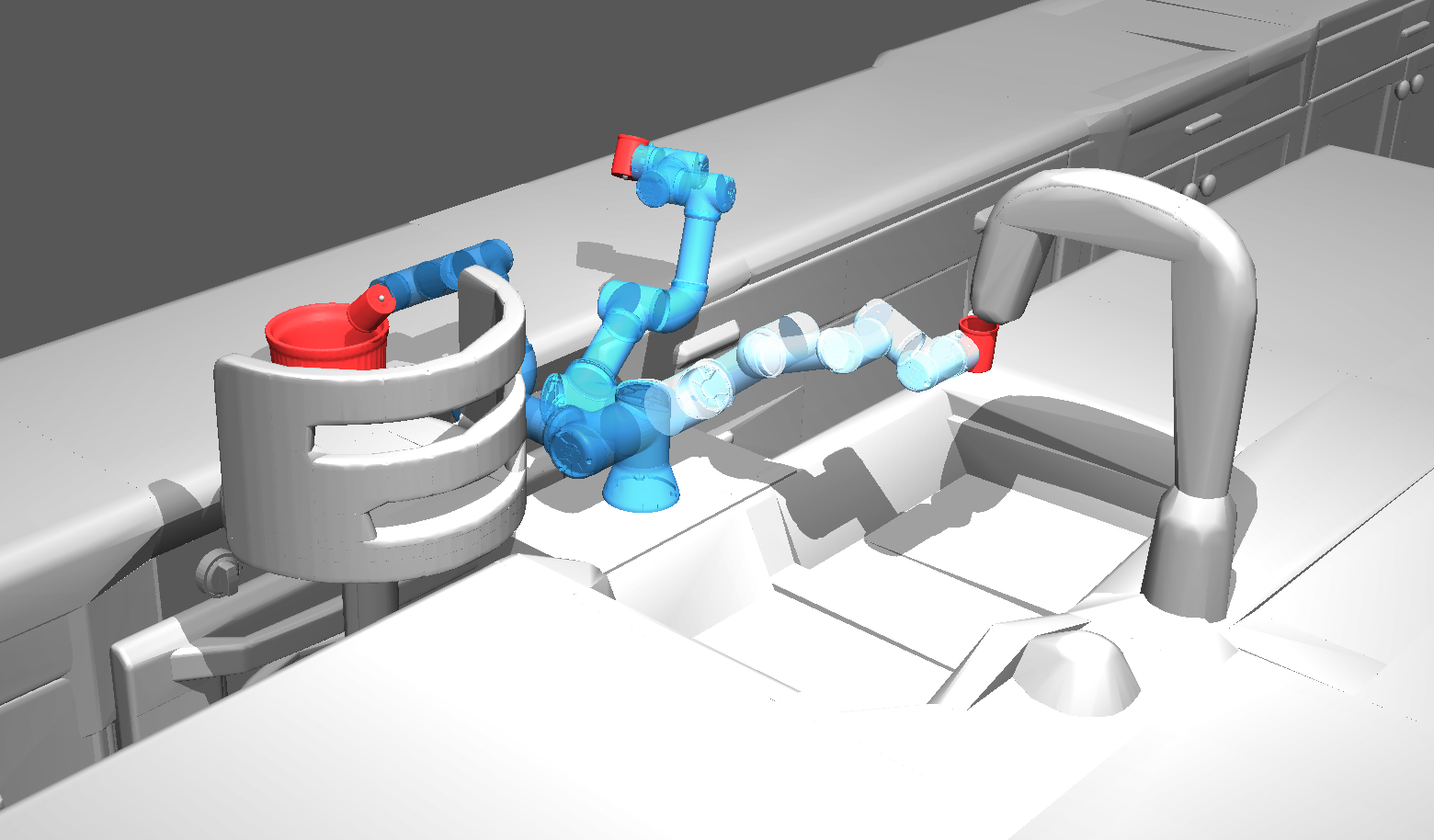}
    \label{subfig:1a}
    }
    \hspace*{1mm}
    \subfloat[]{\includegraphics[height=2.23cm]{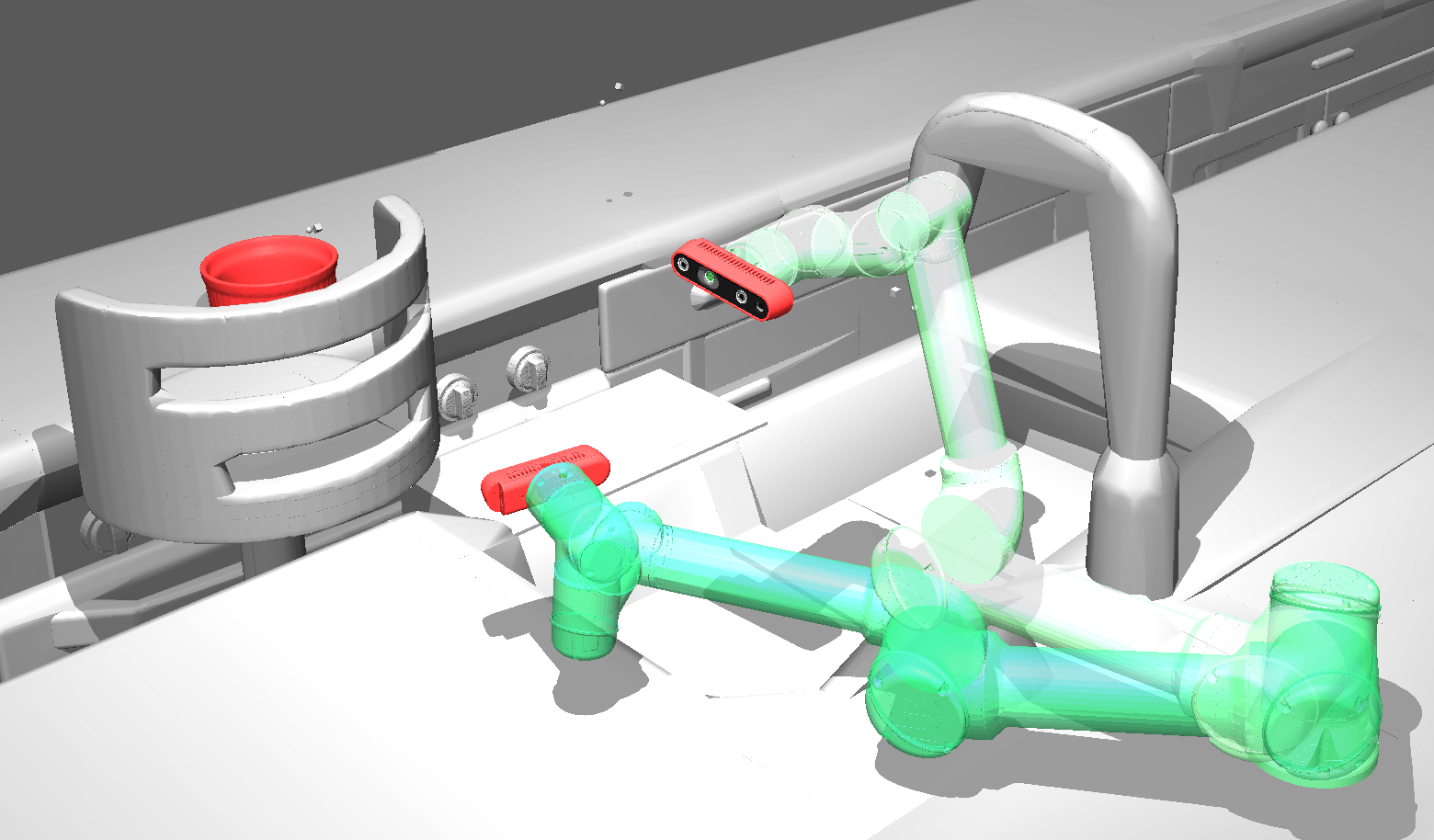}
    \label{subfig:1b}
    }
    \hspace*{1mm}
    \subfloat[]{\includegraphics[height=2.23cm]{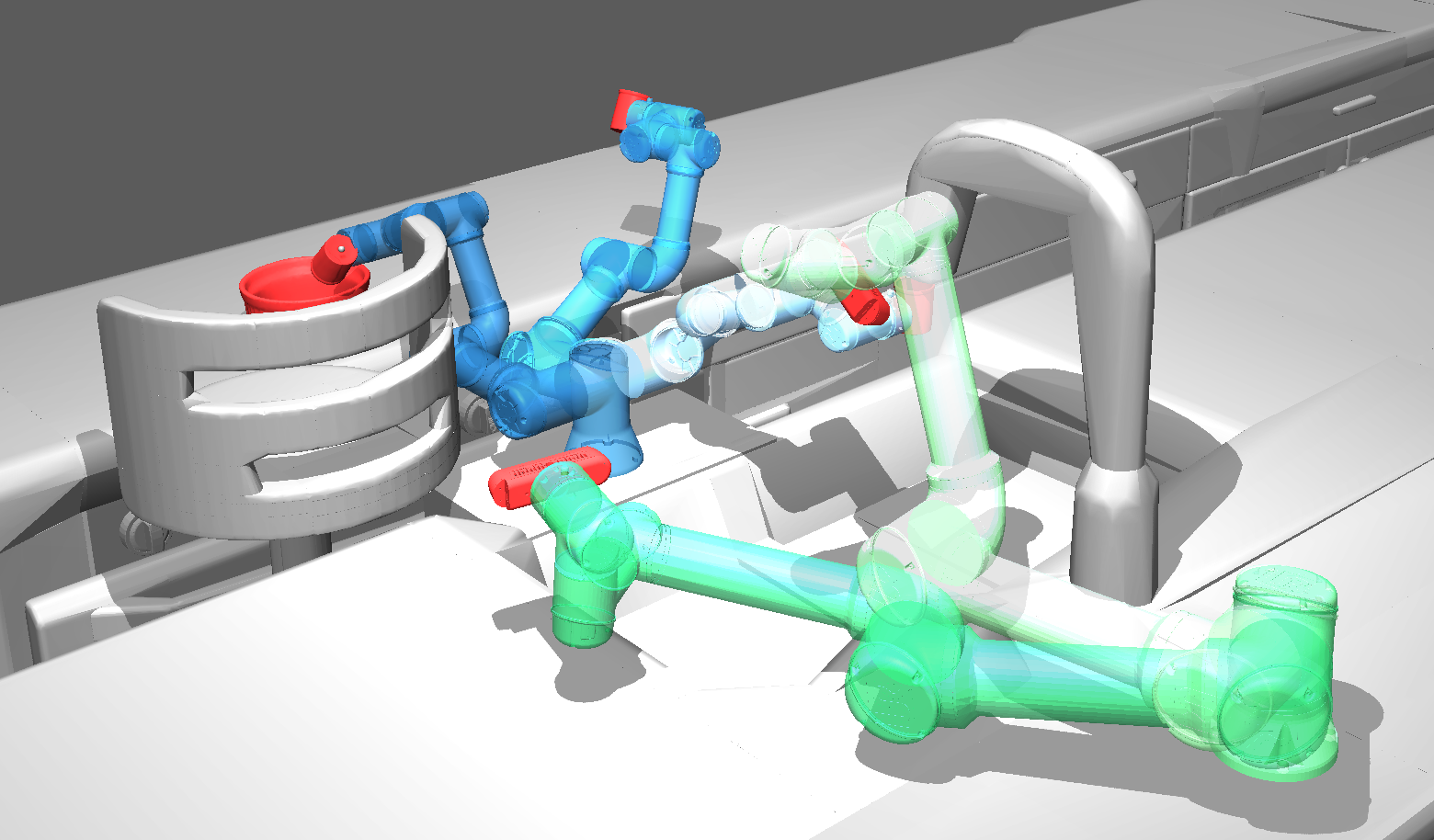}
    \label{subfig:1c}
    }
    \caption{
        %
        TAP in household applications.
        \protect \subref{subfig:1a}~Blue manipulator~$\taskR$ is tasked with transferring water in a cup from a faucet to a pot (light to dark blue correspond with initial to final configurations).
        \protect \subref{subfig:1b}~Green manipulator~$\assistR$ is equiped with a camera that must detect if water is spilled from the cup to initiate clean-up and to ensure that the pot has enough water. 
        Here, the light-green and dark-green robots depict configurations for which the cup is visible and non-visible by~$\assistR$'s point of view, respectively.
        \protect \subref{subfig:1c}~A task assistance path maximizing the amount of time the cup is observed by~$\assistR$ (light to dark green correspond with initial to final configurations). 
        Visualization adapted from~\cite{roberts:2022}.
    }
    \label{fig:motivating-application}
\end{figure}

Examples of assistance include 
visual feedback  and 
communication relays.
For example, visual feedback can be used within the feedback loop of a low-level controller as demonstrated  in Fig.~\ref{fig:motivating-application}. 
Here, a controller is used to  ensure that liquid is not spilled. 
Another such example is semi-autonomous minimally-invasive robotic surgery where~$\taskR$ is a tool tele-operated by a surgeon who is tasked with suturing or removing a tumor and~$\assistR$ is an autonomous endoscope both capable of providing the surgeon with visual feedback by taking a path in which the surgeon’s tool should be as visible as possible as well as controlling the force used by~$\taskR$. 
Alternatively communication relays can be used in search-and-rescue in a limited-communication region: Here,~$\taskR$ is an autonomous ground vehicle (AGV) that needs to communicate with a base in a disaster-ridden area.~$\assistR$ is equipped with a communication-relay device and provides the AGV  a stable communication link by taking a path in which it can retransmit data to the base.

TAP requires solving the motion-planning problem \cite{Hauser2020,L06} where we compute a collision-free path for a robotic system while also accounting for assistance constraints such as visibility constraints \cite{O17}. Unfortunately, the motion-planning problem is already computationally challenging \cite{HSS17,Solovey2020} and adding assistance constraints only further complicates the problem. Roughly speaking, here we need to plan a path for the assistance robot while accounting for when and where assistance is provided. This results in a motion-planning variant where Bellman’s principle of optimality does not hold (i.e., it may be worthwhile not to provide assistance at early stages of the task in order to be able to provide more assistance in later stages of the task). This makes existing motion-planning algorithms unsuitable to address this problem.

A common approach used in MP~\cite{L06,S19} is to 
(i)~create a roadmap~$G$ 
(ii)~solve the original problem restricted to~$G$
and
(iii)~densify~$G$ and repeat step~(ii).
In MP, step (ii) corresponds to solving a shortest-path problem and the literature is abundant with general~\cite{HNR68} 
and application specific~\cite{DellinS16,LimST21,MSS18} 
algorithms that can be used.
In this work we propose to follow a similar approach, however this requires additional care: First, one is required to reason about \emph{timing}: a roadmap needs to be augmented with assistance information. I.e., what part of~$\taskR$'s path can be viewed from each vertex. Second, and more important, is how to solve the TAP problem when restricted to graphs, a problem we dub graph TAP or g-TAP.

Since g-TAP may serve as the basic algorithmic building block to TAP algorithms, in this paper we assume that a roadmap is given (e.g., by running \algname{RRG} or \algname{PRM*}~\cite{KF11}) and restrict our focus to studying g-TAP.
We start (Sec.~\ref{sec:otp}) by considering the most simple setting where we are given the path of~$\assistR$  as a sequence of vertices and only need to decide when it should transition from one vertex to the next. We show that although this is a continuous planning problem, computing transition times that maximize assistance provided can be done in polynomial time.
Moving to the general problem of g-TAP, we start (Sec.~\ref{sec:hardness}) by proving that it is $\NP$-hard\conference{.}{ by a reduction from the subset-sum problem.}
We proceed (Sec.~\ref{sec:optp}) to present a Branch and Bound (B\&B) algorithm that integrates the optimal algorithm  for paths together with a method that allows to efficiently prune large parts of the search space. 
As we demonstrate empirically (Sec.~\ref{sec:experiments}), this algorithm allows to efficiently compute solutions for complex problems significantly outperforming the optimal baseline by roughly three to four orders of magnitude, and successfully computing a solution for significantly larger graphs. Compared to a non-optimal baseline, our algorithm computes paths that can improve assistance  by a factor of roughly $3\times$.
\section{Related Work}
\label{sec:background}

\paragraph{Assisting agents in collaborative settings.}
Our work bares resemblance to research for enabling agents to assess their need for help and their ability to be helpful.
This has been  investigated using the notion of \emph{Value of Information}~\cite{howard1966information,russell1991principles,zilberstein1996intelligent} to quantify the impact information has on autonomous agents’ decisions and utilities. 
However, in contrast to our setting, here there is typically no centralized control, thus requiring coming up with local decisions.
\conference{}{Arguably, the most closely-related work to our new problem is recent work on computing the \emph{Value of Assistance}~\cite{VOI23,VOI23b} which allows to estimate the expected effect an intervention will have on a  robot's belief.
In contrast to our work, this problem is limited to providing assistance at one point along the task-robot's path and the question at hand is where should this assistance be~given.}

Our work also falls under the broad category of multi-robot collaboration~\cite{rizk2019cooperative}. However, here we assume that the path of $\taskR$ is fixed (e.g., when $\taskR$ and $\assistR$ are managed by different systems). We leave the problem of simultaneously planning for $\taskR$ and $\assistR$  to future work.

\paragraph{Visual assistance \& planning with visual constraints.}
Variants of TAP where the assistance is visual feedback have been studied throughout the years but none of the tools developed are directly applicable to our setting.
Specifically, our problem falls under the broad category of robot target detection and tracking which encompasses a variety of decision problems such as coverage, surveillance, and pursuit-evasion~\cite{robin:hal-01183372}.
These kind of problems have typically been studied in the adversarial setting (see, e.g.,~\cite{ChungHI11}) where one group of robots attempts to track down members of another group,
while we are interested in the \myemph{cooperative setting} where the task and  assistance robots work in concert (or at least do not deliberately attempt to jeopardize task assistance).
Moreover, existing work typically considers relatively simple low-dimensional systems (see, e.g.,~\cite{LagunaB19,LaValleGBL97}) in contrast to the high-dimensional ones that we are interested in.

Visual assistance is also closely related to 
planning camera motions (see, e.g.,~\cite{Geraerts09,GoemansO04,NieuwenhuisenO04a,RakitaMG18})  where we are tasked with planning the motions of a free-flying camera to follow a given object.
However, these problems  do not need to account for the robots' potentially high-dimensional configuration spaces and are often studied in relatively uncluttered environments. 
Finally, our problem bares resemblance to the scene-reconstruction problem which has to do with creating a digital model of a real-world scene from a set of images or other measurements of a scene (see, e.g.,~\cite{bircher2016receding,1087372}).
However, in contrast to our problem, here there is no need to account for (i)~time, forcing us to capture images in a pre-defined order and (ii)~collision avoidance, forcing us to account for the geometry of~$\assistR$.

\paragraph{Inspection planning.}
Closely related to our work is the problem of \emph{inspection planning}~\cite{FKSA19,FuSA21}, or \emph{coverage planning}~\cite{almadhoun2016survey,galceran2013RAS}.
Here, we are given a robot \rob with an on-board sensor, a region of interest (ROI)  to be inspected by \rob, and the environment. A point on the ROI is considered inspected if \rob's sensor sees it without any object occluding the view. 
Inspection planning calls for computing a collision-free path for \rob that maximizes the portion of the ROI that is inspected while obeying \rob's kinematic constraints.
%
%
It has been extended to the cooperative setting~\cite{ropero2019terra} but, similar to visual assistance, the order in which POIs are inspected is unimportant which makes algorithms developed for inspection planning difficult to apply to our setting.

\section{Notation \& Problem Definitions}
\label{sec:definitions}
Let $G = (V,E)$ be a graph  which we call a \myemph{task-assistance graph} corresponding to configurations of~$\assistR$. Time is normalized to be in the range $[0,1]$ and each vertex $v \in V$ is associated with \myemph{a set of time intervals}~$\I(v)$ corresponding to the times where assistance can be provided from $v$ (e.g., the times when~$\taskR$'s path can be inspected when the task is visual assistance).\footnote{Note that $\taskR$'s path is not explicitly provided but it is implicitly defined in the task-assistance graph.}
%
%
Additionally, each vertex is associated with a set of valid intervals in which~$\assistR$ is allowed to reside in that vertex (times in which~$\assistR$ can't reside at a vertex allow our model to incorporate avoiding moving obstacles such as $\taskR$).\footnote{To simplify exposition,  unless stated otherwise, we assume that~$\assistR$ is allowed to reside in every vertex during the whole task duration but all of our results can easily be adapted to the general setting.}
Each edge $e \in E$ is associated with a length $\ell(e)$ and we assume for simplicity that 
(i) moving along an edge takes time that is identical to its length and that 
(ii) when moving along an edge $e = (u,v)$, assistance is defined as identical to the assistance at $u$ and at $v$ for the first and second half of the edge~$e$, respectively.\footnote{Our model doesn't account for dynamics such as bounded acceleration but is a sufficient first-order approximation.}

Let $\pi = \path$ be a path such that $v_i \in V$ and $(v_i, v_{i+1}) \in E$. 
As we will shortly see, it will be convenient to introduce several notation:
We set~$\ell_\pi(v_i)$ to be the length of the path from $v_0$ to~$v_i$. 
Namely, $\ell_\pi(v_i):= \sum_{j=0}^{i-1}{\ell(v_j, v_{j+1})}$.
%
Additionally, we set $\ell^-_\pi(v_i)$ and $\ell^+_\pi(v_i)$ to be the length of the path  from $v_0$ to the middle of incoming and outgoing edge of~$v_i$, respectively.
Namely,~$\ell^-_\pi(v_i) := \ell_\pi(v_i)~-~\frac{1}{2}\ell(v_{i-1},v_{i})$
and~$\ell^+_\pi(v_i)~:=~\ell_\pi(v_i)~+~\frac{1}{2}\ell(v_i,v_{i+1})$.
Additionally, set~$\ell^+_\pi(u,v) := \ell^+_\pi(v) - \ell^+_\pi(u)$.
To simplify exposition we define $\ell_\pi(v_{-1}, v_0) := 0$ and $\ell_\pi(v_{k}, v_{k+1}) := 0$.
When understood\conference{}{ from the context,} we omit $\pi$ from $\ell_\pi(v_i), \ell^+_\pi(v_i), \ell^-_\pi(v_i)$ and $ \ell^+_\pi(u, v)$. 
Finally, we denote by $N_{\I}^\pi$ and~$N_{\I}^G$ the total number of intervals of  all vertices in a path $\pi$ and a graph~$G$, respectively.

Importantly, a path only defines where~$\assistR$ is but not when it needs to transition from one vertex to another. Thus, paths need to be augmented  with a sequence of timestamps representing the times at which~$\assistR$ should transition from one vertex to another.
Following our model, these timestamps are defined as the times at which the robot should reach the middle of each edge. 
\begin{dfn}[Timing-profile]
\label{def:timing-profile}
    Let $\pi = \path$ be a path. We define a \myemph{timing-profile} of $\pi$ as a sequence of timestamps~$\T_\pi = \langle t_0, \dots, t_{k-1} \rangle $ such that:
    (i)~$t_0 \geq \ell^+(v_0)$,
    (ii)~$t_{i+1} \geq t_i + \ell^+(v_i, v_{i+1})$
    and
    (iii)~$t_{k-1} + \ell^+(v_{k-1}, v_k) \leq 1$.
    %
    Note that given $\pi$ and $\T_\pi$, it is straightforward to derive the times that~$\assistR$ will arrive and leave each vertex.
\end{dfn}


\begin{figure}[t!]
  \centering
	\subfloat[]{\includegraphics[height=2.5cm]{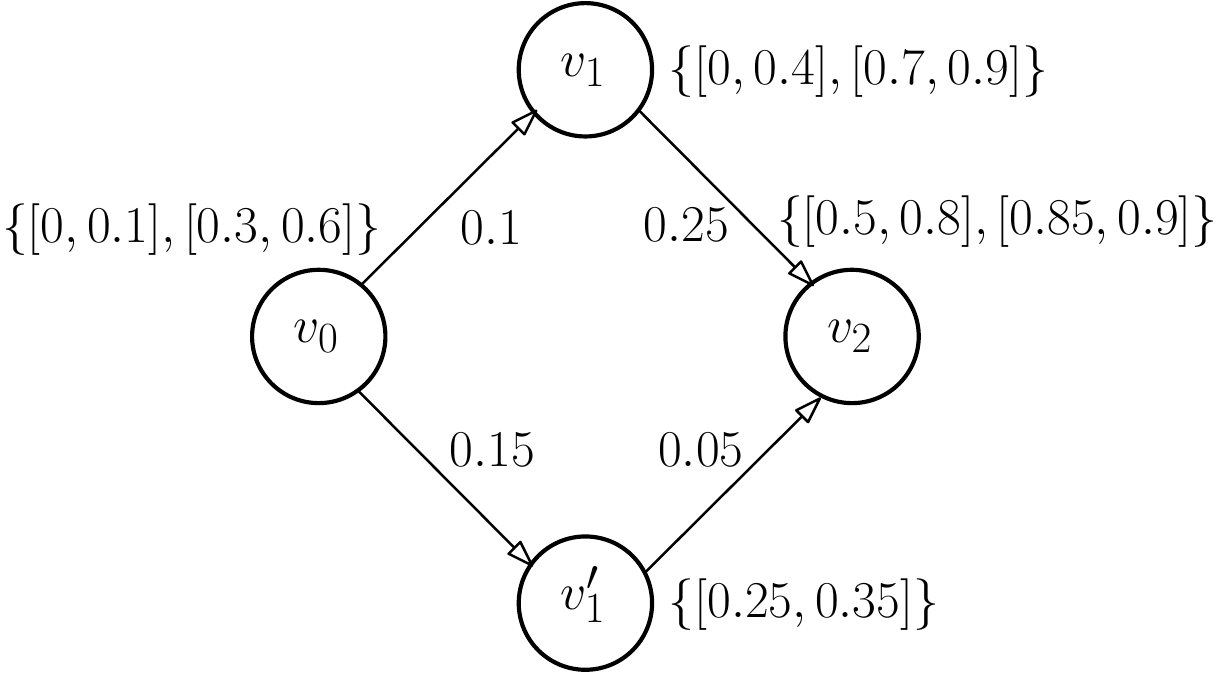}\label{fig:graph}}
	\hfill
	\subfloat[]{\includegraphics[height=2.4cm]{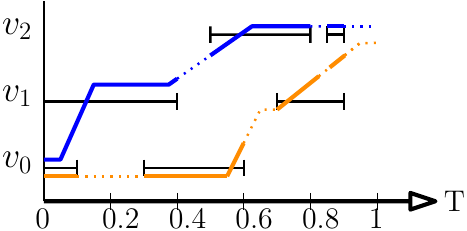}\label{fig:timings}}
 \vspace{-2mm}
  \caption{
 \protect\subref{fig:graph} Toy g-TAP problem.  
  Above each vertex and edge are the intervals for which task assistance can be performed and the edge length, respectively.
  \protect\subref{fig:timings}~Two timing profiles for path $\langle v_0, v_1, v_2 \rangle$. 
  Here, each vertex is depicted together with the intervals for which task assistance can be performed. 
  The timing profiles (blue and orange) consist of solid and dotted lines when task assistance can and can't be performed, respectively.
  Note that the slopes of moving from~$v_0$ to $v_1$ and from~$v_1$ to $v_2$ are different as the corresponding edge lengths are different ($0.1$ and $0.25$, respectively).
  }
 \label{fig:q1}
\vspace{-5mm}
\end{figure}
In order to quantify the effectiveness of the assistance provided, we define the reward as the portion of time where assistance is provided. Formally,
\begin{dfn}[Reward at a vertex]
    Let $u$ be a vertex and let $0 \leq t \leq t'  \leq 1$ be two times, we denote by~$\R(u, t, t')$  the \myemph{reward at vertex $u$} obtained between times $t$ and $t'$. Namely, 
    $\R(u, t, t') = \sum_{I \in \I(u)}{|[t,t'] \cap I|}$.
\end{dfn}

\begin{dfn}[Reward of a timing-profile]
    Let $\pi = \path$ be a path, and let $\T_\pi~=~\tp$ be a timing-profile, 
    we define $\R(\pi, \T_\pi)$ to be the \myemph{reward of the timing-profile $\T_\pi$}.
    Namely, if we set $t_{-1} = 0, t_{k} = 1$, then we have that
    $\R(\pi, \T_\pi) := \sum_{i=0}^{k}{\R(v_i, t_{i-1}, t_i)}$.
\end{dfn}

As an example, consider Fig.~\ref{fig:q1} depicting a g-TAP Instance.
Fig.~\ref{fig:timings} depicts two timing profiles for path $\langle v_0, v_1, v_2 \rangle$: $\T_{\text{blue}} = \langle 0.1, 0.5 \rangle$ and $\T_{\text{orange}}~=~\langle 0.6, 0.825 \rangle$ (recall the times in a timing profile are the times at which the robot reaches the middle of the edges). $\T_{\text{blue}}$ obtains a reward of 
$0.1, 0.3$ and~$0.35$ at $v_1,v_2$ and~$v_3$, respectively. 
Thus, $\R(\pi, \T_{\text{blue}})~=~0.75$ while~$\T_{\text{orange}}$ obtains a reward of~
$0.4, 0.125$ and $0.05$ at $v_1, v_2$ and~$v_3$, respectively. 
Thus, $\R(\pi, \T_{\text{orange}})~=~0.575$.

We are finally ready to define our optimization problems:
\begin{prob}[OTP]
\label{prob:OTP}
    Let $\pi$ be a path and let $\mathcal{T}(\pi)$ be the set of all possible timing profiles over $\pi$. 
    The \myemph{Optimal Timing-Profile (OTP)} problem calls for computing a timing-profile $\T^*$ for $\pi$ whose reward is maximal.
    Namely, compute $\T^*$ s.t.,
    $$
        \T^* \in \argmax_{\T \in \mathcal{T}(\pi)} \R(\pi, \T).
    $$
\end{prob}

\begin{prob}[OPTP]
\label{prob:OPTP}
    Let $G$ be a task assistance graph,
    $v_0 \in V$ a start vertex.
    Let $\Pi(v_0)$ be the set of paths in $G$ starting from $v_0$ and 
    $\mathcal{T}(\pi)$ be the set of all possible timing profiles over a given path~$\pi \in \Pi(v_0)$.
    The \myemph{Optimal Path and Timing-Profile (OPTP)} problem calls for computing a path $\pi^*$ and a timing profile $\T^*$ whose reward is maximal.
    Namely, compute $\pi^*, \T^*$ s.t.,
    $$
    \pi^*, \T^* \in \argmax_{\substack{\pi \in \Pi(v_0), \\ \T \in \mathcal{T}(\pi)}} \R(\pi, \T).
    $$
\end{prob}

\section{Solving the OTP Problem (Prob.~\ref{prob:OTP})}
\label{sec:otp}

\label{subsec:OTP-approach}
At first glance, given a specific path $\pi$, computing an optimal timing-profile $\T$ may seem to be a continuous optimization problem as each timestamp $t \in \T$ can be in the range $[0,1]$. However, our first key insight is that we can consider a discrete set of \myemph{critical times}. 
Roughly speaking, these are  the start and end time of intervals while accounting for edges length. This is because there are two conceptual reasons to leave a vertex~$u$: either an interval $I$ of~$u$ ended and there is no reason to stay at~$u$ (as no additional reward will be gained from $I$), or there is another interval $I'$ at a successor of $u$ we would like to reach further along $\pi$. 

As an example, consider path $\pi=\langle v_0, v_1, v_2 \rangle$ and the timing profile $\T_{\text{blue}}$ (which is optimal) from Fig.~\ref{fig:q1}. Here, $\T_{\text{blue}}$ leaves vertex $v_0$ at time $0.1$ which is exactly the end time of the first interval of $v_0$. Next, $\T_{\text{blue}}$ leaves vertex $v_1$ at time $0.5$ which is exactly the time at which the first interval of $v_2$ starts. As we will show, it is enough to consider only critical times to find an optimal timing profile.
We Formally define critical times in Sec.~\ref{subsec:critical-times} and show in Sec.~\ref{subsec:otp} how they can be used to solve Prob.~\ref{prob:OTP}.

\subsection{Critical Times}
\label{subsec:critical-times}

Given two vertices $v_i,v_j$ in a path $\pi = \path$, 
$v_i \prec_\pi v_j$ denotes that~$v_i$ lies before $v_j$ in $\pi$ (i.e., that $i < j$).
Again, when clear, we omit $\pi$ from~$v_i \prec_\pi v_j$. 
Furthermore, we assume that the first and last vertex in $\pi$ contain the intervals $[\ell^+(v_0), \ell^+(v_0)]$ and $[1 -\ell^+(v_{k-1}, v_{k}), 1 -\ell^+(v_{k-1}, v_{k})]$, respectively\footnote{These represent the earliest time the first vertex can be left and the latest time the last vertex can be reached, respectively. Adding these intervals does not affect the reward as both interval lengths are zero and are only used to simplify the definitions.}.
We start by defining the critical times between two vertices in $\pi$. 
\begin{dfn}[Vertex-pair critical times]
\label{dfn-vpct}
    Let $u, v$ be vertices in path $\pi$ such that $u~\prec~v$. 
    The set of \myemph{vertex-pair critical times}~$\ct(u,v)$ consists of two types of times, defined  as follows:
    \begin{itemize}
        \item[\textbf{T1}] For any interval $I\in  \mathcal{I}(u)$,         
        the earliest time to leave $u$ 
        after $I$ terminates is a type T1 time.
        Namely, all type {T1} times of $\ct(u,v)$ are
        \conference{
            defined as $\bigcup_{ [t_s,t_e] \in  \mathcal{I}(u)} \bigl\{ t_e\bigr\}$.
        }{
        $$
            \bigcup_{ [t_s,t_e] \in  \mathcal{I}(u)} \bigl\{ t_e \bigr\}.
        $$
        }
       
       \item[\textbf{T2}] For any interval $I \in \mathcal{I}(v)$, 
       the latest time needed to leave $u$ in order to reach~$v$ at the start of $I$
       is a type T2 time.
       Namely, all type {T2} times of $\ct(u,v)$ are
        \conference{
            defined as
            $\bigcup_{[t_s,t_e] \in  \mathcal{I}(v)} \bigl\{ t_s - \left(\ell^-(v) - \ell^+(u)\right) \bigr\}$.
        }{
        $$
            \bigcup_{[t_s,t_e] \in  \mathcal{I}(v)} \bigl\{ t_s - \left(\ell^-(v) - \ell^+(u)\right) \bigr\}.
        $$}
    \end{itemize}
\end{dfn}

Each vertex-pair has its own critical times, but the critical times of two pairs sharing a vertex are tightly related. For example, given the path $\langle v_0, v_1, v_2 \rangle$ from Fig.~\ref{fig:graph}, the vertex-pair critical times are:
\begin{align*}
\ct(v_0, v_1) &= \{\textbf{0}, \textbf{0.05}, \textbf{0.1}, \textbf{0.6}, 0.7 \} \\
\ct(v_0, v_2) &= \{\textbf{0}, \textbf{0.05}, \textbf{0.1}, \textbf{0.6}, \underline{0.325}, \underline{0.675} \} \\ 
\ct(v_1, v_2) &= \{0.4, \underline{0.5}, \underline{0.85}, 0.9 \}.
\end{align*}
Notice that \textbf{bold} critical times are identical, and that \underline{underlined} critical times of~$\ct(v_1, v_2)$ are equal to  \underline{underlined} critical times of~$\ct(v_0, v_2)$ shifted by $\ell^+(v_1, v_2)= 0.175$.
We formalize this relation using the following observations.

\begin{obs}
\label{obs:T1}
    Let $u, v, v'$ be vertices in path $\pi$ such that $u \prec v \prec v'$. 
    $t$ is a type T1 in $\ct(u,v)$ 
    iff
    $t$ is a type T1  in $\ct(u,v')$.
\end{obs}

\begin{obs}
\label{obs:T2}
    Let $u, u', v$ be vertices in path $\pi$ such that $u \prec u' \prec v$. 
    $t$ is a type T2 in $\ct(u,v)$
    iff 
    $t + \ell^+(u, u')$ is a type T2 in~$\ct(u',v)$.
\end{obs}
Next, we use the notion of vertex-pair critical times to define vertex-critical times which  include the  times from which $\assistR$  might want to leave a vertex.
\begin{dfn}[Vertex-critical times]
\label{def:ct-i}
    Let $v_i$ be a vertex in path $\pi$.
    The set of \myemph{vertex-critical times}~$\ct_i$ for $v_i$ is defined as follows:
    Let $t_u \in \ct(u,v)$ be a vertex-pair critical time for some vertices $u,v$ in $\pi$ s.t.  $u \prec v$. 
    Then, if
%
    \begin{itemize}
        \setlength{\itemindent}{0.7cm}
        \item[$\mathbf{v_i \prec u}$,]
        $\ct_i$ includes the latest time needed to leave $v_i$ 
        to leave $u$ at time $t_u$.
        \item[$\mathbf{v_i = u}$,]
        $\ct_i$ includes  $t_u$.
        \item[$\mathbf{u \prec v_i}$,]
        $\ct_i$ includes the earliest time we can leave $v_i$ given we left $u$ at time $t_u$.
    \end{itemize}
    Formally,
    $
        \ct_i = \bigcup_{u \prec v}{\{t_u + \ell^+(u, v_i)~|~t_u \in \ct(u,v)\}}.
    $
%
\end{dfn}

Similar to vertex-pair critical times,  vertex-critical times of different vertices are tightly related as well. Following our example from Fig.~\ref{fig:graph} we have that:
\begin{align*}
    \ct_0 &= \{0, 0.05, 0.1, 0.225, 0.325, 0.6, 0.675, 0.7, 0.725 \}, \\
    \ct_1 &= \{0.175, 0.225, 0.275, 0.4, 0.5, 0.775, 0.85, 0.875, 0.9\}, \\
    \ct_2 &= \{0.35, 0.4, 0.525, 0.625, 0.9, 0.975, 1.0, 1.025, 1.35\}.
\end{align*}
Notice that the critical times are identical up to a constant shift.
The following observation formalizes this relation.
\begin{obs}
\label{obs:ct-rel}
    For any vertex $v_i$, the vertex-critical times $\ct_i$ are equal to the vertex-critical times of $\ct_0$ shifted by $\ell^+(v_0, v_i)$. 
    Formally,
    $$\ct_i = \{t + \ell^+(v_0, v_i)~|~t \in \ct_0\}.$$ 
\end{obs}
\noindent

\begin{lem}
    \label{lem:ct-0}
    $\ct_0$ can be computed in $\calO(k + N_\calI^\pi )$ time.
\end{lem}
\begin{proof}[Proof (sketch)]
    Following Obs.~\ref{obs:T1} and~\ref{obs:T2}, there are $\calO( N_\calI^\pi)$ type $T1$ and type $T2$ critical times in $\ct_0$, respectively. 
    Computing these critical times can be done by iterating once over each interval which can be done in $\calO(k + N_\calI^\pi)$ time.
    %
\end{proof}
\noindent
\textbf{Note.} Following Obs.~\ref{obs:ct-rel} we can compute $\ct_i$ in $\calO(\vert \ct_0 \vert)$ given $\ct_0$.

%
%

The next theorem states that an optimal timing-profile can be found by only considering vertex-critical times.
To prove the theorem, we show that any optimal timing profile can be transformed to one that contains vertex-critical times only. See Appendix~\ref{app:otp} for details.

\begin{thm}
\label{thm:otp}
    For any path $\pi =  \langle v_0, \dots, v_{k} \rangle$, there exists an optimal timing profile $\T_\pi = \langle t_0, \dots, t_{k-1} \rangle$ such that~$\forall i: t_i \in \ct_i$.    
\end{thm}



\subsection{Algorithm}
\label{subsec:otp}

\begin{algorithm}[t!]
\caption{OTP}\label{alg:OTP}
\hspace*{\algorithmicindent} 
    \textbf{Input:} path: $\pi = \path$;
    \hspace{3mm} 
    intervals:  $\mathcal{I}$ \\ 
    \hspace*{5mm}
    \textbf{Output:} 
    reward of $\T^*$ : $R_{\text{max}}$
    \cmm{$\T^*$ can be immediately computed}
\begin{algorithmic}[1]

\State  $\entry \gets \{(0, 0)\}$; 
   \hspace{3mm} 
   $\exit \gets \emptyset$   \label{alg1:init}
                   
\For {$v_i \in \pi$} \label{alg1:outer-start}
    \State $R_{\text{max}} \gets 0$
    \For {$t_{\text{exit}} \in \ct_i$ } \cmm{computed using Lemma~\ref{lem:ct-0} and Obs.~\ref{obs:ct-rel}} \label{alg1:inner-start}
        \State $r \gets$ \texttt{compute\_best\_reward}($\entry, v_i, t_{\text{exit}}$) \label{alg1:best-reward} \cmm{Alg.~\ref{alg:cbr}}

        \conference{
        \State $R_{\text{max}} \gets r$
        \State \exit.insert($(t_{\text{exit}}, r)$) \label{alg1:add-to-exit}  }
        {
        \If {$\exit = \emptyset$ or $r > R_{\text{max}}$}  \cmm {Pareto frontier optimization}
        \State $R_{\text{max}} \gets r$ 
        \State \exit.insert($(t_{\text{exit}}, r)$) \label{alg1:add-to-exit}
        \EndIf
        }

    \EndFor \label{alg1:inner-end}
    
    \State $\entry \gets \exit$; 
        \hspace{3mm} 
            $\exit \gets \emptyset$ \label{alg1:next-iter}
\EndFor \label{alg1:outer-end}
\State \textbf{return} $R_{\text{max}}$ \label{alg1:return}

\end{algorithmic}
\end{algorithm}

Thm.~\ref{thm:otp} allows us to present an efficient algorithm that computes the optimal reward (\ie the reward obtained by following an optimal timing-profile). 
Specifically, it implies that there exists an optimal timing profile that belongs to $\ct_0 \times \ldots \times \ct_k$. Clearly, one could iterate over all such  timing profiles but this is highly-inefficient. Instead, we maintain for each vertex a set of so-called \myemph{time-reward pairs}. Such a time reward-pair $(t,r)$ at vertex $v_i$ represent a time~$t \in \ct_i$ and a reward~$r$ that can be obtained by reaching $v_i$ while following a certain timing-profile until time~$t$. These time-reward pairs are computed by iterating along the path vertices one at a time\conference{}{ while removing time-reward pairs that can't belong to an optimal timing profile}.

Our algorithm (Alg.~\ref{alg:OTP}) maintains two time-reward lists $\entry, \exit$ representing the list of optional entry and exit times to a certain vertex, respectively. 
They are initialized to $(0,0)$ (corresponding to the fact that the initial vertex is entered at time zero with zero reward) and an empty list, respectively (Line~\ref{alg1:init}).
The algorithm then proceeds by iterating over the vertices of $\pi$ starting at~$v_0$ (Lines~\ref{alg1:outer-start}-\ref{alg1:outer-end}). 
For each vertex $v_i$, it iterates over all optional exit times~$\texit \in \ct_i$ (Lines~\ref{alg1:inner-start}-\ref{alg1:inner-end}) and for each such exit time~$\texit$, computes the best reward obtainable given that vertex $v_i$ must be left at time $\texit$ (Line~\ref{alg1:best-reward}) using the function \texttt{compute\_best\_reward} (Alg~\ref{alg:cbr}).
This function simply  iterates over all time-reward pairs in $\entry$. For each entry time-reward pair $(t_{\text{entry}}, r_{\text{entry}})$, it computes the reward obtainable by entering vertex~$v_i$ at time~$t_{\text{entry}}$ and leaving it at time~$\texit$ and adds it to the reward obtained prior to entering vertex~$v_i$.
After finishing $v_i$'s iteration, it sets the exit times of $v_i$ as the entry times of~$v_{i+1}$, and sets the exit list to be empty (Line~\ref{alg1:next-iter}).
Finally, the algorithm returns the maximal reward it obtained (Line~\ref{alg1:return}).

\begin{algorithm}[t!]
\caption{\texttt{compute\_best\_reward}($\entry, v_i, t_{\text{exit}}$)}\label{alg:cbr}
\hspace*{\algorithmicindent} 
    \textbf{Input:} entry list of critical times: \entry;
    \hspace{3mm} 
    vertex:  $v_i$;
    \hspace{3mm} 
    exit time:  $t_{\text{exit}}$\\
    \hspace*{5mm}
    \textbf{Output:} 
    best reward given we leave $v_i$ at time $t_{\text{exit}}$ : $R_{\text{best}}$
\begin{algorithmic}[1]

\State $R_{\text{best}} \gets 0$
\For {$(t_{\text{entry}}, r_{\text{entry}}) \in \entry$} \label{alg-cbr:loop-start}
    \If {$t_{\text{entry}} + \ell^+(v_{i-1}, v_{i}) > t_{\text{exit}}$} \label{alg-cbr:if}
    \cmm {$t_{\text{entry}}$ is too late to leave at $\texit$}
        \State \textbf{continue} 
    \EndIf
    
    
    \State $r_{\text{exit}} \gets r_{\text{entry}} + \R(v_i,t_{\text{entry}}, t_{\text{exit}})$  \label{alg-cbr:R-call}
    \hfill //{reward if $v_i$ is visited at $[t_{\text{entry}},t_{\text{exit}}]$}
    \State $R_{\text{best}} \gets \max \{ R_{\text{best}}, r_{\text{exit}}\}$
\EndFor \label{alg-cbr:loop-end}
\State \textbf{return} $R_{\text{best}}$
\end{algorithmic}
\end{algorithm}

\conference{}{
\paragraph{Pareto-frontier optimization.}
Consider the path $\pi = \langle v_0, v_1, v_2 \rangle$ from Fig.~\ref{fig:graph} and recall that~$\ct_0 = \{0, 0.05, 0.1, 0.225, 0.325, 0.6, 0.675, 0.7, 0.725 \}$. After the first iteration of the algorithm, the time-reward pairs of \exit are:
\begin{align*}
\{&{(0.05, 0.05)}, {(0.1, 0.1)}, (0.225, 0.1), {(0.325, 0.125)}, \\ &{(0.6, 0.4)}, (0.675, 0.4), (0.7, 0.4), (0.725, 0.4), (1, 0.4)\}.
\end{align*}
Now, consider the two pairs $p=(0.1, 0.1)$ and $p'=(0.225, 0.1)$. Here,~$p'$ can be pruned as it leaves the vertex after $p$ and its reward is not better.
In the general setting, we only need to maintain the Pareto frontier~\cite{SalzmanF0ZCK23} of time-reward pairs. 
I.e., the set of all time-reward pairs that are not  dominated by any other time-reward pair (a time-reward pair $p_1 = (t_1, r_1)$ dominates a time-reward pair $p_2 = (t_2, r_2)$ if $t_1 \leq t_2$ and $r_1 < r_2$ or if $t_1 < t_2$ and $r_1 \leq r_2$.
%
As the time-reward exit pairs are ordered from earliest to latest, we add a pair $(\texit, r)$ to $\exit$    only if the reward is greater then the reward of the previous pair (Line~\ref{alg1:add-to-exit}).
Returning to our example,  after the first iteration $\exit$ will be pruned down to: 
$\{(0.05, 0.05), {(0.1, 0.1)}, {(0.325, 0.125)}, {(0.6, 0.4)}\}$.
}

\paragraph{Correctness \& complexity (sketch).}

Following Thm.~\ref{thm:otp}, the reward returned by the algorithm corresponds to the reward of an optimal timing-profile. In addition, we can trace back the time-reward pairs in order to find an optimal timing-profile. 

Given a path $\pi$ with $k$ vertices and a total of $N_{\calI}^\pi$ intervals we can bound the size of~$\ct_0$ by $|\ct_0| = \mathcal{O}( N_{\calI}^\pi)$ (Lemma~\ref{lem:ct-0}). In addition, \entry and \exit are both bounded by~$\mathcal{O}(|\ct_0|)$.
Alg.~\ref{alg:cbr} performs $\calO(N_{\calI}^\pi)$ calls of $\R$ (Line~\ref{alg-cbr:R-call}). These calls only differ by their~$t_{\text{entry}}$ value. Since $\entry$ is ordered, the value of $t_{\text{entry}}$ only increases. Thus, by computing the $\R$ value once for the first call, and only substituting the reward for the interval between two following $t_{\text{entry}}$ values, we can compute all $\calO(N_{\calI}^\pi)$ calls performed in Alg.~\ref{alg:cbr} in $\calO(N_{\calI}^\pi)$ time.

To summarize, for each of the $k$ vertices, we call Alg.~\ref{alg:cbr} $\mathcal{O}(N_{\calI}^\pi)$ times, and each calls takes~$\calO(N_{\calI}^\pi)$ time.
Thus, the total runtime complexity 
is $\mathcal{O}\left(k \cdot \left( N_{\calI}^\pi \right)^2\right)$.

%
\section{Hardness of the OPTP Problem}
\label{sec:hardness}
We now move on to the OPTP problem and show that it is $\NP$-hard.
Roughly speaking, the hardness of the problem comes from the fact that in the general setting, there may be an exponential number of paths in a graph and in contrast to the shortest-path problem, Bellman’s principle of optimality does not hold. 
%

\begin{thm}
    \label{thm:hardness}
    The OPTP problem is $\NP$-hard.
\end{thm}

\begin{proof}[Sketch]
    The proof is by a reduction from the subset-sum problem (SSP)~\cite{kleinberg2006algorithm}.
    Recall that the SSP is a decision problem where we are given 
    a set 
    $X = \{x_1, \dots, x_n \mid~x_i \in \mathbb{N} \}$
    and a target 
    $\k \in \mathbb{N}^{+}$\conference{}{(for simplicity, we assume that $\k>0$ but this is a technicality only)}. 
    The problem calls for deciding whether there exists a set $X' \subseteq X$ whose sum $\sum_{x\in X'} x$~equals $K$.

    Given an SSP instance, we build a corresponding OPTP instance (to be explained shortly) and show that there exists a subset $X' \subseteq X$ such that $\sum_{x\in X'} x = \k$ iff the optimal reward in our OPTP instance is $\frac{\k}{\sum_{i=1}^{n}{x_i}} + 1$ thus proving the problem is $\NP$-hard.
    
    W.l.o.g. assume that 
    $x_1, \dots, x_n$ are sorted from smallest to largest
    and set
    $\kappa_i:= \sum_{j=1}^{i} x_j$,
    $\alpha_i := x_i / \kappa_n$ and $y_i = \sum_{j=1}^{i-1}{\kappa_j}$. 
    The graph of our OPTP instance depicted in Fig.~\ref{fig:reduction} contains 
    $n$ hexagons $H_1, \ldots H_n$ such that $H_i$ contains vertices~$a_i, b_i, c_i, d_i, e_i, f_i$. 
    We call $a_i$ and $f_i$ the entry and exit vertices of $H_i$,
    $b_i$ and $c_i$ the top of $H_i$
    and
    $d_i$ and $e_i$ the bottom of $H_i$.
    Edges $(b_i,c_i)$ and $(d_i,e_i)$ at the top and bottom of $H_i$ have length $\kappa_i$ and $\kappa_{i-1}$, respectively (all other edge lengths are equal to zero).
    Edge $(b_i,c_i)$ at the top of $H_i$ contains the interval $[y_i, y_i + \alpha_i]$.\footnote{Strictly speaking, the interval belongs to the vertices $b_i, c_i$. However, it will be more convenient to consider the interval as belonging to the edge $(b_i,c_i)$.} 
    Finally, the exit $f_n$ of the last hexagon $H_n$ has an edge to a vertex~$u$ which has a valid interval $[0, y_n + \k]$ and $u$ has an edge to a vertex~$v$ whose assistance interval is $[y_n + \k, y_n + \k + 1]$.
    Note that here for ease of exposition, time is in the range $[0, y_n+\k+1]$ and \emph{not}  $[0,1]$.

\begin{figure}[tb]
    \centering
    \includegraphics[scale=1]{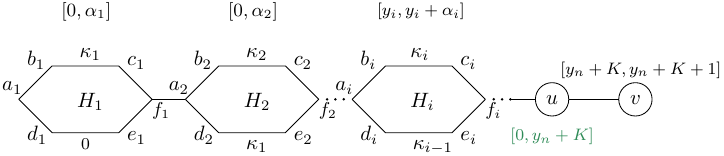}
    \caption{
        Reduction graph (all edges are directed from left to right). 
        When omitted, edge length equals zero.
    }
    \label{fig:reduction}
    \vspace{-5mm}
\end{figure}

     Our reduction is based on two key properties of the new OPTP instance:
    \begin{itemize}
        \item[P1]
        The shortest path to reach $u$ is by taking the lower part of each $H_i$ and its length is $y_n$.
        \item[P2] 
        Let $\pi$ be a path that passes through $u$.
        Going through the upper part of~$H_i$ adds additional $x_i$ time to reach $u$ and results in an additional reward of~$\alpha_i$.
    \end{itemize}

    Roughly speaking, the valid interval at~$u$ forces any path $\pi$ to $v$ to leave~$u$ before time $y_n + \k$. 
    As the minimal time to reach $u$ is $y_n$ (Property~$P1$),~$\pi$ only has $\k$ time units to spend on earning rewards from the hexagons.
    Now, to obtain a reward of $\k/\kappa_n + 1$,
    $\pi$  must earn a reward of $\k/\kappa_n$
    before reaching~$u$.
    Since the upper part of $H_i$ adds an additional time of $x_i$ and a reward of $\alpha_i$ (Property~$P2$),~$\pi$ must find a combination of upper parts $I_{\text{up}}$ such that $\sum_{i \in I_{\text{up}}}{x_i} = \k$ which is the exact solution to the subset-sum problem.

    For additional details see Appendix~\ref{app:hardness}.
\end{proof}

\section{Solving the OPTP Problem (Prob.~\ref{prob:OPTP})}
\label{sec:optp}
Given a task assistance graph $G = (V,E)$ and a start vertex $v_0$, we can iterate over all paths starting from $v_0$, and for each path run the OTP algorithm (Sec.~\ref{sec:otp}). Unfortunately, the search space can be extremely large (Sec.~\ref{sec:hardness}) which deems this approach intractable.

To this end, we suggest to apply the Branch and Bound (B\&B) framework~\cite{Clausen2003BranchAB} to our setting. 
B\&B is a general algorithmic technique used to solve optimization problems, particularly combinatorial-optimization problems, by systematically exploring the solution space in a structured manner. 
Conceptually, B\&B divides the solution space into smaller subspaces (branching) and then systematically searches through these subspaces while keeping track of bounds on the optimal solution (bounding). This allows  to prune branches of the search tree that are guaranteed not to contain an optimal solution, thereby reducing the size of the search space and improving efficiency.

Thus, we start (Sec.~\ref{subsec:optp-bounds}) by introducing approaches to bound the reward of partial solutions and then continue (Sec.~\ref{subsec:optp-bnb}) to detail our B\&B-based algorithm.

\vspace{-2.5mm}
\subsection{Upper Bound}
\label{subsec:optp-bounds}
In the following, we overview how to bound
the reward obtainable from 
(i)~any path that starts at the beginning of an interval~$I$,
(ii)~any path that starts at a vertex $u$ starting at time $t$
and
(iii)~from a prefix of a path $\pi$.
Here, we give a high-level description regarding how these bounds are computed and refer the reader to Appendix~\ref{app:bound} for additional details.

\vspace{-2.5mm}
\subsubsection{Bounding the reward obtainable from the beginning of an interval}
Let~$I = [t_s, t_e]$ be an interval belonging to vertex $u$.
We set $\ub_I^0 = \vert I \vert$ and iteratively compute~$\ub_I^i$ using $\ub_I^{i-1}$. As we will see,  $\ub_I^i$ will be an upper bound on the reward that may be obtained from interval $I$ followed by at most~$i$ subsequent intervals.
Note that~this indeed holds for $i=0$ and
that if the invariant holds for every iteration, then after $N_{\calI}^G$ iterations
$\ub_I:=\ub_I^{N_{\calI}^G}$ is an upper bound on the total reward obtainable starting from interval $I$ and continuing optimally.
To compute~$\ub_I^i$, we iterate over all intervals $I'$,
and bound the reward obtainable assuming the interval after $I$ is $I'$. 
%
This is done by computing $t_{\text{unused}}$, which is the minimal portion of either $I$ or $I'$ in which no reward can obtained when traveling from $I$ to $I'$.
We then bound the reward as $\vert I \vert + \ub^{i-1}_{I'} - t_{\text{unused}}$.
%
%
If the bound obtained is greater than $\ub_I^i$, we update it accordingly.

Importantly, $\ub_I^i$ stores an upper bound on the reward obtained from interval~$I$ as well as from future intervals. However, it does \emph{not} store the time $I$ is exited in order to obtain the reward from future intervals. Namely, $\ub_I^i$ does not account for the fact that $I$ may have been entered at a time $t > t_s$.
This results in (i)~an upper bound on the true reward and (ii)~an efficient algorithm whose complexity is cubic in $N_{\calI}^G$ (the number of intervals) and not on the number of critical times which may be exponential in~$N_{\calI}^G$.
Note that this process is computed \emph{once}, before running our B\&B-based algorithm and will be used to compute the other bounds required by our algorithm.
%

Before stating the correctness of $\ub_I$, we  introduce the following definition: 
\begin{dfn}[Partial reward]
    Let $\pi$ be a path, let $\T$ be a timing-profile and let $t \in [0,1]$.
    We define $\R(\pi, \T, t)$ to be the reward obtained from following the timing-profile $\T$ during the time interval $[t, 1]$.
\end{dfn}
%
%
\begin{lem}
\label{lem:ub-i}
    Let $\pi$ and $\T$ be a path and timing-profile, respectively. 
    Let $I=[t_s,t_e]$ be an interval~$\T$ obtains reward from. Then it holds that $\ub_I \geq \R(\pi, \T, t_s)$.
\end{lem}

\subsubsection{Bounding the reward obtainable from a vertex $u$ starting at time $t$}
Let $u \in V$ be a vertex and $t \in [0, 1]$. 
Denote $\delta'(u,v)$ to be the minimum distance between~$u$ and $v$ in $G$ while not accounting for half of the first and last edge (this can be computed using one Dijkstra-like pass over the graph together with some additional processing). 
Now, consider an interval $I \in \calI(v)$ associated with some vertex $v$. 
Intuitively, 
(i)~the reward that $I$ offers (assuming we start at vertex~$u$ at time~$t$) cannot be obtained before time~$t + \delta'(u,v)$
and
(ii)~$\ub_I$ bounds the reward that can be obtained from $I$.
Together, these allow to bound the reward obtainable from $u$ at $t$ assuming $I$ is the next interval.
We iterate over all such intervals and set $\ub(u,t)$ to be the highest reward.

%
\begin{lem}
\label{lem:ub-ut}
    Let $u \in V$ be a vertex, and $t \in [0,1]$. 
    For every path $\pi$ and timing-profile $\T$
    s.t. $u \in \pi$ and that following $\T$ implies that at time $t$ the assistance robot is at vertex $u$, it holds that $\ub(u, t) \geq \R(\pi, \T, t)$. 
\end{lem}

\subsubsection{Bounding the reward obtainable given the prefix of a path}
Given a path~$\pi= \langle v_0, \dots, v_k \rangle$, we wish to compute an upper bound on the reward that can be obtained from any path $\pi' = \langle v_0, \dots, v_k, \dots, v_m \rangle$ whose prefix is $\pi$.

Let $\T' = \langle t'_0, \dots, t'_{m-1} \rangle$ be an optimal timing profile for $\pi'$ (note that we don't have access to $\pi'$ and $\T'$ but we will address this shortly).
Furthermore, let $I = [t_s, t_e] \in \calI(v_i)$ for some $v_i \in \pi$ be the last interval that $\T'$ obtained reward from before leaving $v_k$.
We bound $\pi'$'s reward by separating it into two parts: the reward obtained
(i)~until $t'_i$ (the time $\T'$ exits $v_i$)
and 
(ii)~from~$t'_i$.

To bound the reward obtained until $t'_i$ (first part), recall that the OTP algorithm keeps track of time-reward pairs representing a time and the best reward that can be obtained until that time at a certain vertex.
As $I$ is the last interval in $\pi$ that $\T'$ obtains reward from, the reward obtained until $t'_i$ can be bounded by the reward obtained until $t_e$. This is exactly the reward of the pair $(t_e, r)$ belonging to the \exit list of vertex $v_i$ which  can be computed by running the OTP algorithm on $\pi$.
To bound the reward obtained from $t'_i$ (second part) we make use of $\ub(u,t)$. 
Recall that we do not actually have access to $t'_i$ but, since~$\T$ obtains reward from $I$ it must leave $v_i$ after $t_s$, which allows us to bound the reward obtainable from~$t'_i$ by $\ub(v_i, t_s)$. 
%
%
%
%
%
In addition, since $\T'$ does not obtain any reward between the end of $I$ and the time it leaves $v_k$, we can further tighten our bound to $\ub \left( v_{k}, t_s + \ell^+(v_i, v_{k}) \right)$.
Combining both parts, $\R(\pi',\T')$ can be bounded by $r + \ub \left( v_{k}, t_s + \ell^+(v_i, v_{k}) \right)$.

As we don't have access to $\pi'$, we can't know which interval $I$ is the last interval $\T'$ obtains reward from. 
Thus, we must compute this bound for every interval $I$ on the path $\pi$ and use the maximal bound obtained to be~$\ub(\pi)$\conference{.}{, 
our bound on the reward obtainable from the prefix $\pi$ of a path.}
\begin{thm}
\label{thm:ub-pi}
    Let $\pi = \path$ be a path. For any path $\pi' = \langle v_0, \dots, v_k, \dots, v_m \rangle$ extending $\pi$ s.t., $\T'_{\pi'} = \langle t'_0, \dots, t'_{m-1} \rangle$ is its optimal timing profile, it holds that   $\ub(\pi) \geq \R(\pi', \T'_{\pi'})$.
\end{thm}


\paragraph{Complexity analysis}
Given a graph with $n$ vertices, $m$ edges, and $N_{\calI}^G$ intervals,
computing~$\ub_I$ requires computing $\delta'(u,v)$ for every two vertices $u,v \in V$ which can be done in $\calO \left( n \cdot m \cdot \log(n) \right)$. In addition, each iteration $i$ used to compute~$\ub_I^i$ iterates over all pairs of $I, I'$. Since there are~$N_{\calI}^G$ iterations this takes $\calO( (N_{\calI}^G)^3)$. Thus, computing $\ub_I$ can be done in $\calO \left( n \cdot m \cdot \log(n) + (N_{\calI}^G)^3 \right)$. 
Computing $\ub(u,t)$ requires iterating once over each interval thus taking~$\calO(N_{\calI}^G)$ time.
Finally, given a path $\pi$ with $k$ vertices and~$N_{\calI}^\pi$ intervals, computing $\ub(\pi)$ requires running the OTP algorithm over $\pi$, and calling $\ub(u,t)$ once for every interval. Thus, computing $\ub(\pi)$ can be done in~$\calO(k \cdot (N_{\calI}^\pi)^3 + N_{\calI}^\pi \cdot N_{\calI}^G)$.

\subsection{Branch and Bound}
\label{subsec:optp-bnb}

We are finally ready to describe our B\&B-based algorithm (Alg.~\ref{alg:b&b}).
This recursive algorithm is given a path $\pi = \path$ (initialized to the start vertex~$v_0$) and returns the maximal reward obtainable by any path whose prefix is $\pi$. 
The algorithm starts by checking if the path can be traversed in $t<1$ time (Line~\ref{alg4:start}).
If so, it runs the OTP algorithm to compute $\pi$'s optimal reward (Line~\ref{alg4:otp}). 
Subsequently, it checks if any sub-path appended to $\pi$ can improve the currently-stored best reward~$R_{\text{max}}$. 
This is done by checking if $\ub(\pi)$ is greater than $R_{\text{max}}$ (Line~\ref{alg4:ub-check}). 
If this is the case, the algorithm iterates over all adjacent vertices to~$\pi$'s  last vertex (Lines~\ref{alg4:loop-start}-\ref{alg4:loop-end}). For every such vertex $v_{k+1}$ the algorithm appends $v_{k+1}$ to $\pi$ (Line~\ref{alg4:append}), and performs a recursive call to compute the maximal reward obtainable from any path extending the new path (Line~\ref{alg4:recursive}). 
It then updates the maximal reward if needed (Line~\ref{alg4:update}). 
Finally, it returns the overall  maximal reward obtained (Line~\ref{alg4:ret}). 

\begin{algorithm}[t]
\caption{Branch and Bound (B\&B)}\label{alg:b&b}
\hspace*{\algorithmicindent} 
\textbf{Input:} 
    graph: $G = (V,E)$; \hspace{3mm} 
    intervals:  $\mathcal{I}$  \\ 
    \hspace*{1.6cm} 
    path: $\pi = \path$ ;   \cmm {initialized to $\pi \leftarrow \langle v_0 \rangle$}\\
    \hspace*{1.6cm}
    reward: $R_{\text{max}}$ \cmm {initialized to $R_{\text{max}} \leftarrow  0$}\\
    \hspace*{5mm}
    \textbf{Output:} 
    reward of optimal path and timing profile extending $\pi$.
\begin{algorithmic}[1]

\If {$ \ell^+(v_k) > 1$}  \cmm {minimal time to reach the end of $\pi$}\label{alg4:start}
    \State \textbf{return} $0$  \cmm {last vertex is not reachable}
\EndIf

\vspace{2mm}

\State $R_{\text{max}} \gets \max\{R_{\text{max}}, \text{\textbf{OTP}}(\pi, \I)\}$  \cmm {run OTP}\label{alg4:otp}

\vspace{2mm}

\If {$\ub(\pi) \leq R_{\text{max}}$} \label{alg4:ub-check}
        \State \textbf{return} $R_{\text{max}}$ \cmm {prune all search space containing paths appended to $\pi$}
\EndIf

\vspace{2mm}

\For{\textbf{each} $v_{k+1}$ s.t. $(v_k, v_{k+1}) \in E$} \label{alg4:loop-start}
    \State $\pi' \gets \langle v_0, \dots, v_k, v_{k+1} \rangle$ \label{alg4:append}
    \State $R \gets \text{B\&B}(G, \calI, \pi', R_{\text{max}})$ \cmm {recursive call} \label{alg4:recursive}
    \State $R_{\text{max}} \gets \max\{R_{\text{max}}, R\}$  \label{alg4:update}
\EndFor \label{alg4:loop-end}

\State \textbf{return} $R_{\text{max}}$ \label{alg4:ret}

\end{algorithmic}
\end{algorithm}

%
\begin{thm}
    Given
    a task-assistance graph $G=(V,E)$, an interval mapping~$\calI$ and a vertex $v_0 \in V$,
    Alg~\ref{alg:b&b} solves the corresponding OPTP problem (Prob.~\ref{prob:OPTP}).
    %
\end{thm}

To improve the algorithm's  runtime, we apply several optimizations:

\conference{}{
\paragraph{Cycle pruning.} 
In contrast to the shortest-path problem, cycles may be beneficial in our setting: They allow to return to previously-visited vertices in future timesteps to make use of multiple intervals associated with the same vertex.
However, if a path ends by a cycle which does not provide additional reward, we prune all paths extending this path. This can be done during the algorithm in $\calO(1)$ time using some additional bookkeeping. 
}

\paragraph{Interval splitting.}
%
Let $\pi = \path$ be a path, $\T^*$ its optimal timing profile and $I=[t_s, t_e]$ the last interval $\T^*$ obtains reward from before $v_k$. Let $v_i$ be the vertex $I$ belongs to, one can show that $\ub(\pi)$ counts $I$ twice since given the pair $(t_e, r)$ from the \exit list of $v_i$, it upper bounds the reward obtainable from that pair using $\ub(v_i, t_s)$ instead of $\ub(v_i, t_e)$ which allows for $I$ to be counted once in the reward $r$ and once in $\ub(v_i, t_s)$.
Thus, the smaller the intervals are, the tighter $\ub(\pi)$ is.
Thus, we introduce a parameter~$\delta_{\max}$.
Before running our algorithm, we split every interval of size larger than $\delta_{\max}$ to a sequence of  intervals, each of size less than $\delta_{\max}$.
Note that 
(i)~as $\delta_{\max}$ decreases, $\ub(\pi)$ is tighter but 
the number of intervals increases which causes a cubic increase in the complexity of the OTP algorithm (see  Sec.~\ref{subsec:otp}) which is used when computing $\ub(\pi)$
and
(ii)~this optimization does not affect the solution to the OPTP problem.

\conference{}{
\paragraph{Caching \& filtering upper bounds.}
The computational bottleneck for computing~$\ub(\pi)$ for some path $\pi$ is the frequent calls to~$\ub(u,t)$
(once for every interval in the path). 
Moreover, for every $\Delta t >0$ it holds that $\ub(u,t) \geq \ub(u,t + \Delta t) $.
Thus, every time $\ub(\pi,t)$ is computed for some $\pi$ and  $t$, this value is stored.
If, in some subsequent iteration, the algorithms requires to test if the value 
$\ub(u,t + \Delta t) < \tau$ for some  $\Delta t >0$ and some $\tau$, we first test if 
$\ub(u,t) < \tau$.
Only if this is not the case, we compute and store $\ub(u,t + \Delta t)$.
}

\paragraph{Bounded sub-optimality}
We introduce a second hyper-parameter $\eps \geq 0$ and replace 
the condition 
$\ub(\pi) \leq R_{\max}$
in Line~\ref{alg4:ub-check}
with 
$\ub(\pi) \leq (1+\eps) \cdot R_{\max}$.
This allows to dramatically prune the search space and one can easily show that 
if~$R_\eps$ and $R^*$ are the rewards obtained with this variation and the optimal reward, respectively, then $R_\eps \geq R^* / (1+\eps)$.

\vspace{-1.5mm}
\section{Empirical Evaluation}
\label{sec:experiments}
\vspace{-1.5mm}

To evaluate our B\&B-based algorithm (Alg.~\ref{alg:b&b}) we consider the TAP-problem of visual assistance. Specifically, the assistance robot is equipped with a camera that should keep the end-effector of the task robot as-visible as possible.

\begin{figure}[t!]
  \centering
	\subfloat[]{\includegraphics[height=2.75cm]{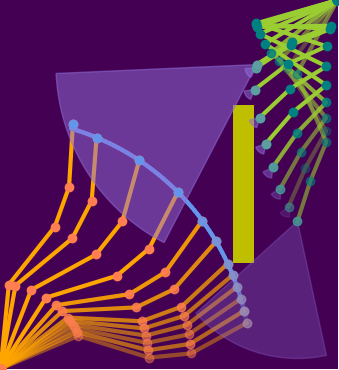}\label{fig:two-obs}}
	\hspace{8.5mm}
	\subfloat[]{\includegraphics[height=2.75cm]{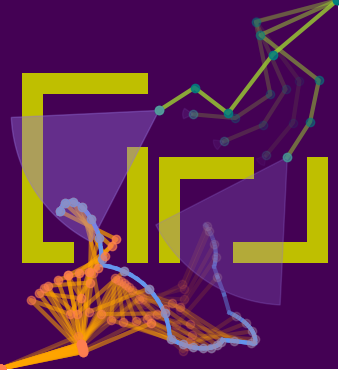}\label{fig:two-rooms}}
 	\hspace{8.5mm}
	\subfloat[]{\includegraphics[height=2.75cm]{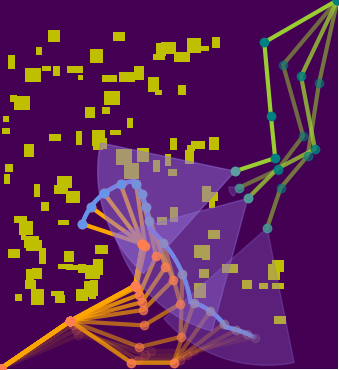}\label{fig:random}}
  \caption{
  Simulated environments consisting of a task-robot (orange),
  assistance-robot (green) and 
  obstacles (yellow).
  The task-robot end-effector follows a predefined path (blue)
  and needs to be located in the field of view of a limited-range camera located on the assistance robot's end effector (purple).
  }
 \label{fig:exp1}
\end{figure}

We consider 
a simulated environment in which both $\assistR$ and $\taskR$  are four-link planar manipulators (Fig.~\ref{fig:exp1}) 
as well as the setting depicted in Fig.~\ref{fig:motivating-application} wherein $\assistR$ is a UR5 and $\taskR$ is a UR3~\cite{UR}.
In each scenario, we fix the path of $\taskR$ and generate a roadmap $G$ using the \rrg algorithm~\cite{KF11} containing between $10$ to $1,000$ vertices ($G$  was stored with increments of ten vertices).  
Time intervals in $G$ are computed by computing the visible region from each configuration and testing what times the path of  $\taskR$  intersects this region.

\conference{We}{As we optimize our B\&B-based algorithm by employing interval splitting
and by allowing for bounded sub-optimality, we} 
present our B\&B algorithm with two parameters indicating the interval size~$\delta_{\max}$ used for interval splitting and the approximation factor $\eps$.
\conference{When}{As we run our algorithm on graphs generated by adding additional vertices and edges, when} running the algorithm on a graph with $n$ vertices, we use the reward obtained from the previous iteration (on the graph with  $n - 10$ vertices) to initialize $R_{\max}$. 
This allows our algorithm to get close to zero runtimes on iterations where the added vertices can not increase the maximal reward.
Consequentially, we present accumulated runtimes.

As  baselines to compare with, we suggest the following strawman algorithms:
The first, which we term ``$\delta$-discretization'' (DD($\delta$)) discretizes the time into steps of size $\delta$. It runs a best-first search where nodes are pairs consisting  of a vertex and a time with the initial node being $\langle v_0, 0\rangle$ (i.e., the start vertex and time $t=0$). When expanding a node $\langle u, t\rangle$ it can either stay at $u$ for~$\delta$ time (resulting in the node $\langle u, t + \delta \rangle$) or move to a vertex $v$ such that $(u,v) \in E$. Note that this algorithm does not guarantee to compute optimal solutions.
The second algorithm, which we term DFS-OTP iterates over all possible paths in the graph in a DFS-like approach (stopping when the length of the path exceeds~$1$). When reaching the final vertex of a path, it runs the OTP algorithm on the path. 
Note that this algorithm is optimal and is in fact  identical to the B\&B approach with a trivial bound of $1$ 
as it iterates over all paths in the graph.

All algorithms were implemented in \Cpp and run on a Linux, x86\_64 server using an Intel Xeon CPU at 2.10GHz with a timeout of one hour.
Code will be made publicly available upon paper acceptance.
\begin{figure}[t!]
   \centering
    \includegraphics[width=12cm]{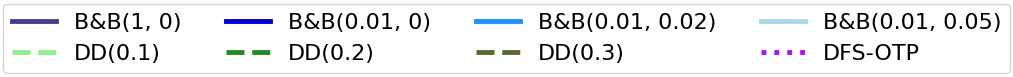}

	\subfloat[]{\includegraphics[width=2.84cm]{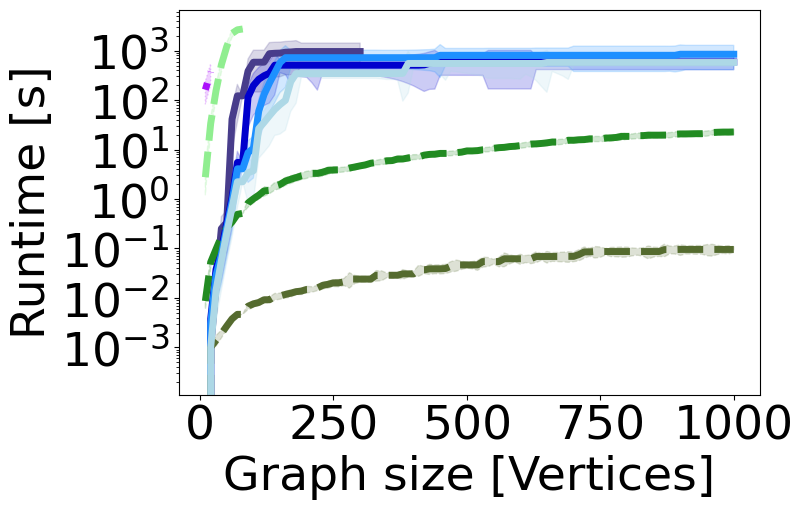}\label{fig:rt1}}
	\hspace{1.5mm}
	\subfloat[]{\includegraphics[width=2.84cm]{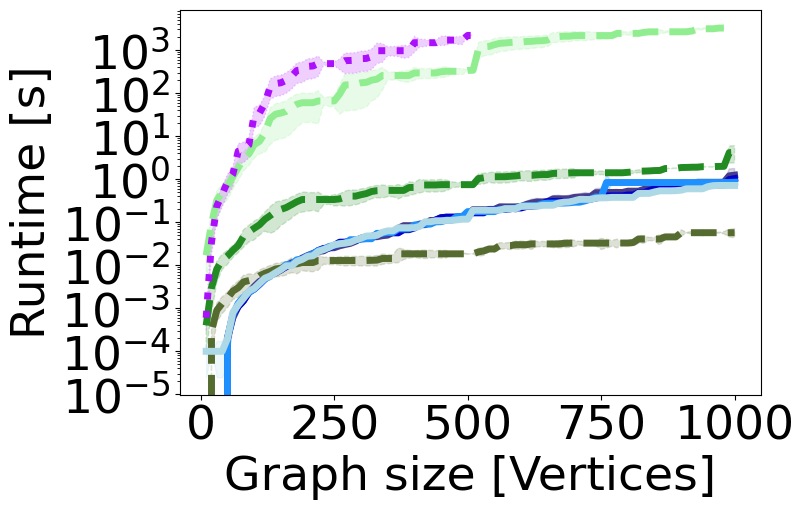}\label{fig:rt2}}
    \hspace{1.5mm}
    \subfloat[]{\includegraphics[width=2.84cm]{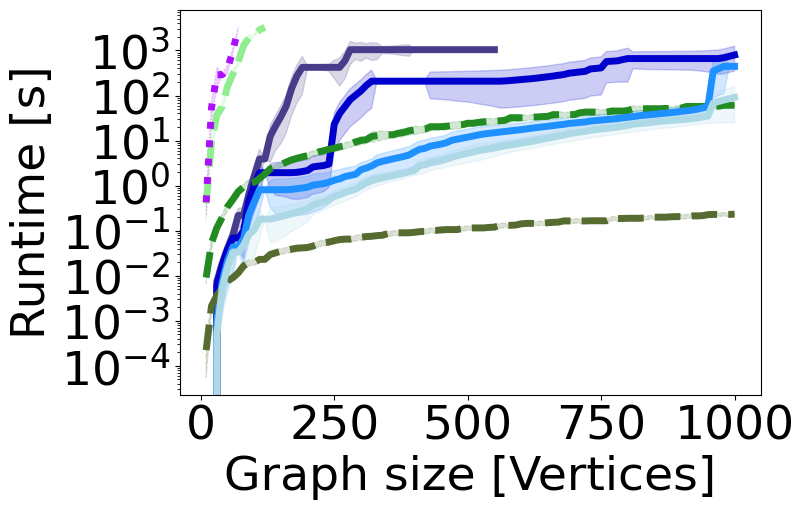}\label{fig:rt3}}
    \hspace{1.5mm}
    \subfloat[]{\includegraphics[width=2.84cm]{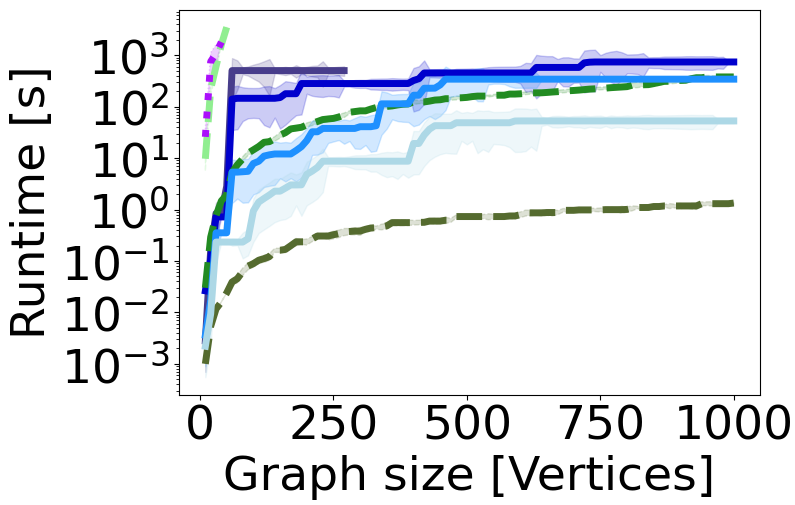}\label{fig:rt4}}

    \subfloat[]{\includegraphics[width=2.84cm]{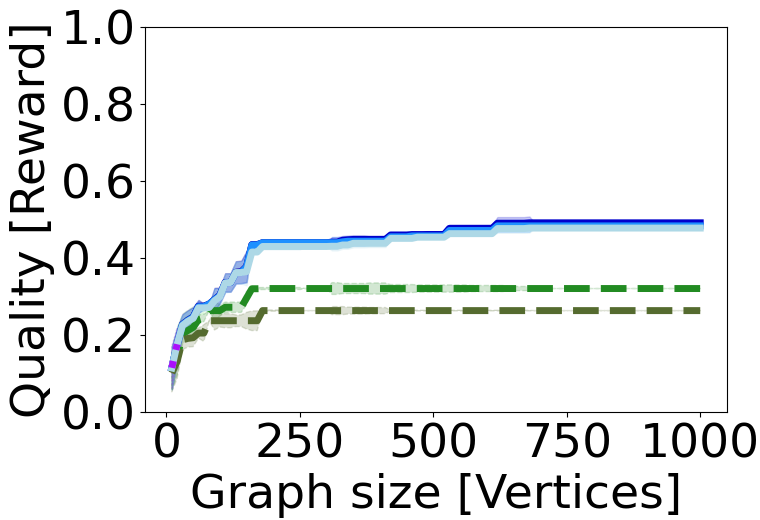}\label{fig:rw1}}
	\hspace{1.5mm}
	\subfloat[]{\includegraphics[width=2.84cm]{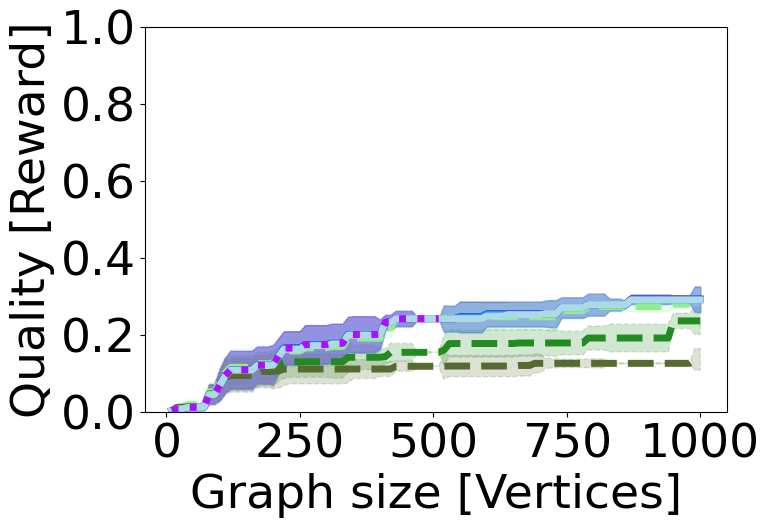}\label{fig:rw2}}
    \hspace{1.5mm}
    \subfloat[]{\includegraphics[width=2.84cm]{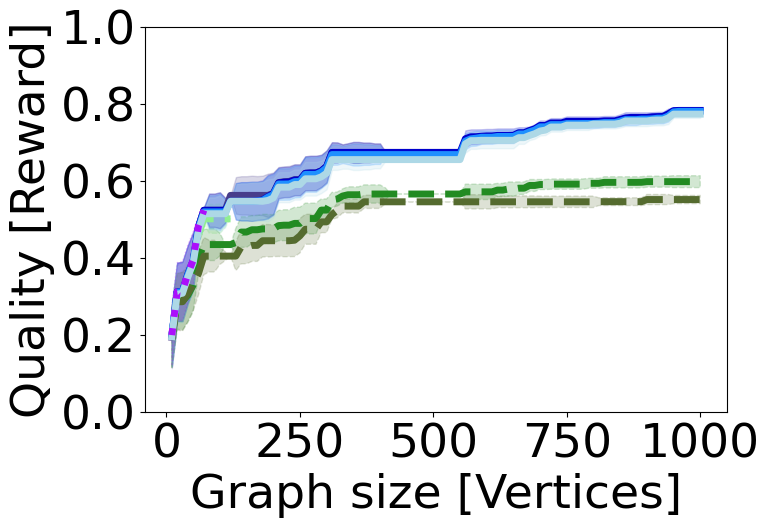}\label{fig:rw3}}
    \hspace{1.5mm}
    \subfloat[]{\includegraphics[width=2.84cm]{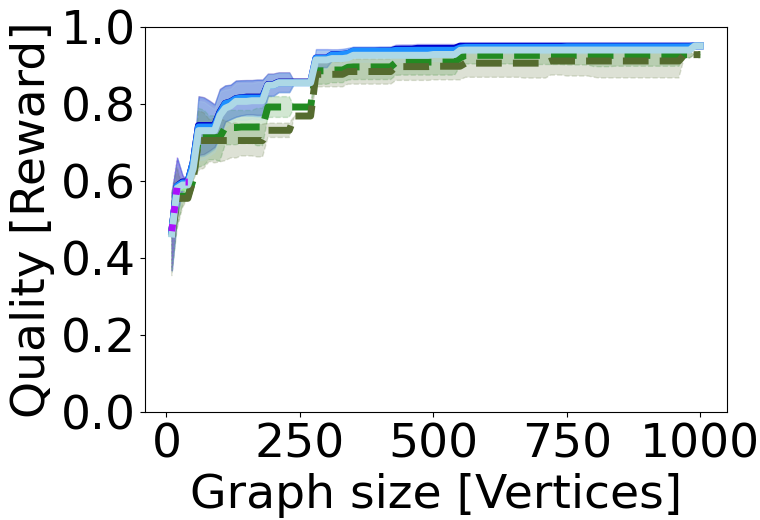}\label{fig:rw4}}
   \caption{
    Running time and quality of solution as a function of graph size for variants of our algorithm and the two baselines.
    Here, plots
    \protect \subref{fig:rt1}-\protect \subref{fig:rt3} 
    and
    \protect \subref{fig:rw1}-\protect \subref{fig:rw3} 
    correspond to the environments depicted in Fig.~\ref{fig:two-obs}-\ref{fig:random}, respectively
    while \protect \subref{fig:rt4} and~\protect \subref{fig:rw4} correspond to the environment depicted in Fig.~\ref{fig:motivating-application}.
    Results are averaged over ten roadmaps with the shaded part corresponding to one standard deviation.  
    }
 \label{fig:runtimes}
 \vspace{-5mm}
\end{figure}

For each one of the four scenarios (three planar manipulators in Fig.~\ref{fig:exp1} and the UR from Fig.~\ref{fig:motivating-application}),  we fixed the path of the task robot and generated ten roadmaps.
We report in Fig.~\ref{fig:runtimes} the average running time and the reward  for each algorithm as a function of the number of graph vertices.
We present four versions of our B\&B algorithm, one with $\delta_{\max} = 1, \eps = 0$ (\ie an optimal algorithm without interval splitting) and three with the same value for $\delta_{\max}$ (for interval splitting) but different values of $\eps$ (the approximation factor)
as well as DD($\delta$)  with three different values of  $\delta$ (smaller $\delta$ values are expected to run longer but obtain higher-quality results) and the DFS-OTP algorithm.

When compared to the baseline optimal algorithm (DFS-OPT), our algorithm  allows for an improved runtime by roughly three orders of magnitude for a given graph size and allows to compute solutions for far larger graphs within the allotted planning time of one hour.
When compared to the baseline heuristic algorithm (DD($\delta$)), while being slower, our algorithm obtains higher-quality results by a factor of up to $3\times$ on the three planar scenarios. For the URs, the problem is much easier containing fewer intervals, thus both algorithms produce comparable results (though ours provides guarantees on the solution quality).

As depicted in Fig.~\ref{fig:runtimes}, $\delta_{\max}$ and $\eps$  can have a large effect on the performance of our B\&B algorithm. Roughly speaking, $\delta_{\max}$ balances how much time each OTP call takes versus how tight the upper bound is, and $\eps$ reduces the search space size.
Recall (Sec.~\ref{sec:optp}), that when computing $\ub(\pi)$, the main gap between the real reward obtainable and the upper bound comes from allowing one interval to be counted twice. Thus, smaller intervals result in a smaller gap.
However, $\delta_{\max}$ results in more intervals which increases the runtime of the OTP algorithm. 
If the approximation factor $\eps$ is larger than the aforementioned gap, it will allow the algorithm to explore only paths whose quality is significantly better that the incumbent solution.

Finally, for a video of UR robots running our algorithm on a scenario similar to Fig.~\ref{fig:motivating-application}, see \url{https://tinyurl.com/yhhurh4n}.
\section{Future Work}
\label{sec:discussion}
In this work we lay the algorithmic foundation for TAP.
Here, we assumed that 
(\textbf{A1})~the roadmap $G$ is provided 
and that
(\textbf{A2})~the path of $\assistR$ is known.
In future work we wish to relax both assumptions.
To relax \textbf{A1} we intend to build an \rrt-like algorithm that reasons about where to sample while taking the task timing into account.
To relax \textbf{A2} we suggest to estimate the path of $\assistR$ using some learning algorithm and then iteratively run our B\&B algorithm  in a model-predictive control (MPC) fashion.
Finally, it may be interesting to consider the \emph{joint}  problem of simultaneously planning the path of  $\taskR$ and $\assistR$.

\conference{}{
\section*{Acknowledgments}
We wish to thank 
Shaull Almagor for his assistance in proving the hardness of the OTPT problem (Sec.~\ref{sec:hardness})
and 
Dan Elbaz, Ofek Gottlieb for their assistance in the empirical evaluation.
}

\bibliographystyle{splncs04}
\bibliography{tip_ast, mypub-os}
\vfill

\pagebreak
\setcounter{page}{1}
\appendix
\section{OPT---Proof of Thm.~\ref{thm:otp}
}
\label{app:otp}
\newcommand{\opt}{^*}

\begin{reptheorem}{thm:otp}
 For any path $\pi =  \langle v_0, \dots, v_{k} \rangle$, there exists an optimal timing profile $\T_\pi = \langle t_0, \dots, t_{k-1} \rangle$ such that~$\forall i: t_i \in \ct_i$. 
\end{reptheorem}

\begin{proof}
To prove the theorem, we will show that any optimal timing profile can be transformed to one that contains vertex-critical times only. 

Let $\T^* = \langle t\opt_0, \dots, t\opt_{k-1} \rangle$ be an optimal timing-profile for $\pi$, and let $i$ be the first index such that $t\opt_i \notin \ct_i$.
We describe a procedure in which we create another timing-profile $\T' = \langle t'_0, \dots, t'_{k-1}
\rangle$ such that 
(i)~$\R(\pi, \T') \geq \R(\pi, \T^*)$
and that
(ii)~$t'_j = t\opt_j$ for all $j < i$ and $t'_i \in \ct_i$. 
Consequently, we can take an optimal timing-profile and run this procedure until an optimal  timing-profile exists s.t.  $\forall i~t\opt_i \in \ct_i$ (this takes at most $k$ iterations). 

Our procedure considers two cases: 
Either there is or there isn't an interval $I \in \mathcal{I}(v_i)$ such that $t\opt_i \in I$. 

\vspace{2mm}
\noindent
\textbf{Case 1:}
There is an interval $I \in \mathcal{I}(v_i)$ such that $t\opt_i \in I$.

Here, we consider two different sub-cases corresponding to whether $I$ is the last interval that $\T^*$ obtains reward from or not.

\begin{itemize}
    \item[\textbf{1.1:}] $I$ is the last interval that $\T^*$ obtains reward from.

    Since $I$ is the last interval $\T^*$ obtains reward from, $\T'$ can leave later without losing any reward. 
    Recall (Sec.~\ref{sec:otp})
    the last vertex on the path contains the interval $[1 -\ell^+(v_{k-1}, v_{k}), 1 -\ell^+(v_{k-1}, v_{k})]$, thus $1 -\ell^+(v_{k-1}, v_{k}) \in \ct_k$. From Def.~\ref{def:timing-profile}, it holds that $t\opt_k \leq 1 -\ell^+(v_{k-1}, v_{k})$, thus, by setting $t'_j := 1 -\ell^+(v_{j}, v_{k})$ for every $j \geq i$ we get that $t'_j \in \ct_j$ (Obs.~\ref{obs:T2}) and $t'_j \geq t\opt_j$. Hence, if we define $\T' = \langle t\opt_0, \dots t\opt_{i-1}, t'_i, \dots, t'_k \rangle$ it holds that:
\[
\R(\pi, \T') - \R(\pi, \T^*) = \R(\pi, \T', t\opt_i) - \R(\pi, \T^*, t\opt_i) = \R(\pi, \T', t\opt_i) - 0 \geq 0.
\]

    \item[\textbf{1.2:}] $I$ is not the last interval that $\T^*$ obtains reward from.

    Let $I = [t_s, t_e] \in \mathcal{I}(v_i)$ be an interval such that $t\opt_i \in I$, and $I$ is not the last interval~$\T^*$ obtains reward from. Let $I' =[t'_s, t'_e] \in~\calI(v_j)$ be the next interval~$\T^*$ obtains reward from after $I$. Note that $v_i \prec v_j$.
    %
    Here we distinguish between the setting where $t\opt_{j-1} \in (t_s, t_e]$ and $t\opt_{j-1} \notin (t_s, t_e]$:

    \begin{itemize}
        \item[\textbf{1.2.1:}] $t\opt_{j-1} \in (t_s, t_e]$.

        Here, we will  change $t'_i$ to leave vertex $v_i$ earlier than $t\opt_i$. Intuitively, $\T'$ will lose some reward from $I$ but, it will earn the same reward back from $I'$. 
        First, we show that $t\opt_{j-1} = t\opt_i + \ell^+(v_i, v_{j-1})$. Assume by contradiction that $t\opt_{j-1} \neq t\opt_i + \ell^+(v_i, v_{j-1})$, thus $t\opt_{j-1} > t\opt_i + \ell^+(v_i, v_{j-1})$. Set $t'_m := t'_i + \ell^+(v_i, v_m)$ for $i < m \leq {j-1}$. It holds that:
\begin{align*}
    \R(\pi, \T') - \R(\pi, \T^*) &= 
        \sum_{m=i}^{j+1}{\R(v_m, t'_{m-1}, t'_m)} - \sum_{m=i}^{j+1}{\R(v_m, t\opt_{m-1}, t\opt_m)} \\
        &= 0 + \R(v_j, t'_{j-1}, t\opt_j) - (0 + \R(v_j, t\opt_{j-1}, t\opt_j)) \\
        &= \R(v_j, t'_{j-1}, t\opt_{j-1}) + \R(v_j, t\opt_{j-1}, t\opt_j) - \R(v_j, t\opt_{j-1}, t\opt_j) \\
        &= \R(v_j, t'_{j-1}, t\opt_{j-1}) \\
        &= t\opt_{j-1} - \max\{t'_{j-1}, s\} > 0.
\end{align*}
Which means that $\R(\pi, \T') > \R(\pi, \T^*)$ which contradicts the optimality of $\T^*$. Thus indeed, $t\opt_{j-1} = t\opt_i + \ell^+(v_i, v_{j-1})$.

Next, we will look at $\hat{t}_1 := t\opt_{i-1} + \ell^+(v_{i-1}, v_i)$, which is the earliest time $\T'$ can leave $v_i$, and at $\hat{t}_2 := t'_s - \ell^+(v_i, v_{j-1})$, which is the latest time $\T'$ can leave $v_i$ to reach $v_j$ at time $t'_s$. Set $t'_i := \max\{\hat{t}_1, \hat{t}_2\}$, from Obs.~\ref{obs:T1} and~\ref{obs:T2} it holds that $t'_i \in \ct_i$, and since $t\opt_{j-1} = t\opt_i + \ell^+(v_i, v_{j-1})$ and $t\opt_j > t_s$ it holds that $t'_i \leq t\opt_i$.
Now set $t'_m := t'_i + \ell^+(v_i, v_m)$ for $i < m \leq {j-1}$, and $\T' = \langle t\opt_1, \dots t\opt_{i-1}, t'_i, \dots t'_{j-1}, t\opt_j, \dots, t\opt_k \rangle$. It holds that:
\vspace{-2.5mm}
\begin{align*}
    \R(\pi, \T') - \R(\pi, \T^*) &= 
        \sum_{m=i}^{j+1}{\R(v_m, t'_{m-1}, t'_m)} - \sum_{m=i}^{j+1}{\R(v_m, t\opt_{m-1}, t\opt_m)} \\
        &= \R(v_i, t'_{i-1}, t'_i) + \R(v_j, t'_{j-1}, t'_j) - \R(v_i, t\opt_{i-1}, t\opt_i) - \R(v_j, t\opt_{j-1}, t\opt_j) \\ 
        &= -\R(v_i, t'_i, t\opt_i) + \R(v_j, t'_{j-1}, t\opt_{j-1}) \\
        &= -(t\opt_i - t'_i) + t\opt_{j-1} - t'_{j-1} \\
        &= t'_i - t\opt_i + t\opt_i + \ell^+(v_i, v_{j-1}) - t'_i - \ell^+(v_i, v_{j-1}) = 0
\end{align*}

\vspace{-2.5mm}
\item[\textbf{1.2.2:}] $t\opt_{j-1} \notin (t_s, t_e]$.

Since $\T'$ obtains reward from $I'$, it must hold that $t\opt_{j-1} \leq t_s$. We  look at $\hat{t}_1 := t_s - \ell^+(v_i, v_{j-1})$ which is the latest time $\T'$ can leave $v_i$ to arrive to $v_j$ at time $t'_s$, and $\hat{t}_2 := t\opt_j - \ell^+(v_i, v_j)$, which is the latest time~$\T'$ can leave $v_i$ to be able to leave $v_j$ at time $t\opt_j$. 
Set $t'_i := \min\{\hat{t}_1, \hat{t}_2\}$, from Obs.~\ref{obs:T2} it holds that $t'_i \in \ct_i$ and from its definition, it holds that $t'_i \geq t\opt_i$.
Set $t'_m := t'_i + \ell^+(v_i, v_m)$ for $i < m \leq {j-1}$, and $\T' = \langle t\opt_1, \dots t\opt_{i-1}, t'_i, \dots t'_{j-1}, t\opt_j, \dots, t\opt_k \rangle$. It holds that:
\vspace{-2.5mm}
\begin{align*}
    \R(\pi, \T') - \R(\pi, \T^*) &= 
        \sum_{m=i}^{j+1}{\R(v_m, t'_{m-1}, t'_m)} - \sum_{m=i}^{j+1}{\R(v_m, t\opt_{m-1}, t\opt_m)} \\
    &= \R(v_i, t'_{i-1}, t'_i) - \R(v_i, t\opt_{i-1}, t\opt_i) \\
    &= \R(v_i, t\opt_{i-1}, t\opt_i) + \R(v_i, t\opt_{i}, t'_i) - \R(v_i, t\opt_{i-1}, t\opt_i) \\
    &= \R(v_i, t\opt_{i}, t'_i) \geq 0
\end{align*}
        
    \end{itemize}

\end{itemize}

\vspace{2mm}
\noindent
\textbf{Case 2:}
There is no interval $I \in \mathcal{I}(v_i)$ such that $t\opt_i \in I$.

Since $t\opt_i$ is not part of an interval, $\T'$ can leave vertex $v_i$ earlier without losing any reward. The two times that are of interest are the earliest time we can leave $v_i$ given we entered at time $t\opt_{i-1}$, and the end time of the last interval of $v_i$ before $t\opt_i$.
By definition, the earliest time $\T'$ can leave vertex $v_i$ is $\hat{t}_1 := t\opt_{i-1} + \ell^+(v_{i-1}, v_i)$. Since $\hat{t}_1$ is the earliest time $\T'$ can leave $v_i$ it holds that $\hat{t}_1 \leq t\opt_i$ and since $t\opt_{i-1} \in \ct_{i-1}$ it holds that $t'_i \in \ct_i$ (Obs.~\ref{obs:T2}). 
If $v_i$ has at least one interval that ends before $t\opt_i$ starts, we set $\hat{t}_2$ to be the end time of the last interval ending before $t\opt_i$. It holds that $\hat{t}_2 \leq t\opt_i$ and from Def.~\ref{def:ct-i} it holds that $\hat{t}_2 \in \ct_i$. If $v_i$ has no such interval we set $\hat{t}_2 := \hat{t}_1$.
We set $t' := \max\{\hat{t}_1, \hat{t}_2\}$ and $\T' = \langle t\opt_0, \dots, t\opt_{i-1}, t', t\opt_{i+1}, \dots, t\opt_{k-1} \rangle$. 
It holds that:
\[
\R(\pi, \T') - \R(\pi, \T^*) = \R(v_{i+1}, t', t\opt_i) - \R(v_i, t', t\opt_i) = \R(v_{i+1}, t', t\opt_i) - 0 \geq 0.
\]
%
\end{proof}

\section{OPTP Hardness---Proof of Thm.~\ref{thm:hardness}}
\label{app:hardness}

\begin{reptheorem}{thm:hardness}
    The OPTP problem is $\NP$-hard.
\end{reptheorem}

    The proof is by a reduction from the subset-sum problem (SSP)~\cite{kleinberg2006algorithm}.
    Recall that the SSP is a decision problem where we are given 
    a set 
    $X = \{x_1, \dots, x_n \mid x_i \in \mathbb{N} \}$
    and a target 
    $\k \in \mathbb{N}^{+}$ (for simplicity, we assume that $\k>0$ but this is a technicality only). 
    The problem calls for deciding whether there exists a set $X' \subseteq X$ whose sum $\sum_{x\in X'} x$~equals $K$.

    Given an SSP instance, we build a corresponding OPTP instance (to be explained shortly) and show that there exists a subset $X' \subseteq X$ such that $\sum_{x\in X'} x = \k$ iff the optimal reward in our OPTP instance is $\frac{\k}{\sum_{i=1}^{n}{x_i}} + 1$ thus proving the problem is $\NP$-hard.
    
    W.l.o.g. assume that 
    $x_1, \dots, x_n$ are sorted from smallest to greatest
    and set
    $\kappa_i:= \sum_{j=1}^{i} x_j$.
    $\alpha_i := x_i / \kappa_n$ and $y_i = \sum_{j=1}^{i-1}{\kappa_j}$. 
    The graph of our OPTP instance depicted in Fig.~\ref{fig:reduction2} contains 
    $n$ hexagons $H_1, \ldots H_n$ such that $H_i$ contains vertices~$a_i, b_i, c_i, d_i, e_i, f_i$.
    We call $a_i$ and $f_i$ the entry and exit vertices of $H_i$,
    $b_i$ and~$c_i$ the top of $H_i$
    and
    $d_i$ and $e_i$ the bottom of $H_i$.
    Edges $(b_i,c_i)$ and $(d_i,e_i)$ at the top and bottom of $H_i$ have length $\kappa_i$ and $\kappa_{i-1}$, respectively (all other edge lengths are equal to zero).
    Vertices $b_i,c_i$ at the top of $H_i$ contains the interval $[y_i, y_i + \alpha_i]$.
    Finally, the exit $f_n$ of the last hexagon $H_n$ has an edge to a vertex~$u$ which has a valid interval $[0, y_n + \k]$ and $u$ has an edge to a vertex~$v$ whose assistance interval is $[y_n + \k, y_n + \k + 1]$.
    Note that here for ease of exposition, time is in the range $[0, y_n+\k+1]$ and \emph{not}  $[0,1]$. 
    
\addtocounter{figure}{-3}
\begin{figure}[b]
    \centering
    \includegraphics[scale=1]{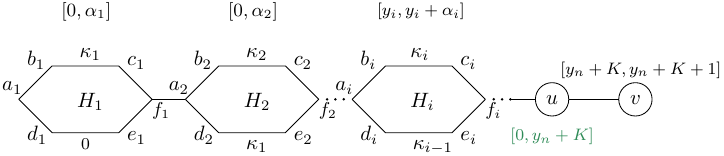}
    \caption{
        Reduction graph (all edges are directed from left to right). 
        When omitted, edge length equals zero.
        (This figure is identical to Fig.~\ref{fig:reduction} in Sec.~\ref{sec:hardness} and is added to make the appendix self-contained).
    }
    \label{fig:reduction2}
\end{figure}
\addtocounter{figure}{3}

Before we prove Thm.~\ref{thm:hardness}, we state several observations and Lemmas.

\begin{obs}
\label{obs:no-wait}
    Let $\pi = \path$ be a path and $\T = \langle t_0, \dots, t_{k-1} \rangle$ a timing profile. If $t_i = t_{i-1} + \ell^+(v_{i-1}, v_i)$ then $\pi$ does not wait at any vertex (except maybe the last one).
\end{obs}

\begin{obs}
\label{obs:down-path}
    By choosing to go through the lower part of all hexagons (which is the fastest way to reach $u$) a timing profile can reach u at time $y_n$.
\end{obs} 
To see why Obs.~\ref{obs:down-path} holds, 
let $\T = \langle t_0, t_1, \dots, t_{k-1} \rangle$ such that $t_0 = 0$, and $t_i = t_{i-1} + \ell^+(v_{i-1}, v_i)$. 
Now we have that indeed,
    \begin{align*}
        t_{k-1} &= t_0 + \sum_{i=1}^{k-1}{\ell^+(v_{i-1}, v_i)} 
        = \sum_{i=1}^{n-1}{\kappa_i} = \sum_{i=1}^{n-1}{\sum_{j=1}^{i}{x_j}} = y_n.
    \end{align*}

\begin{obs}
\label{obs:choosing-up}
    Let $\pi$ be a path, for every hexagon $H_i$ where $\pi$ goes through the upper part, it adds an additional $x_i$ to the time it takes to reach $u$.
\end{obs}

\begin{lem}
\label{lma:reward}
    Let $\pi = \path$ be a path and $\T = \langle t_0, \dots, t_{k-1} \rangle$ a timing profile such that $t_i = t_{i-1} + \ell^+(v_{i-1}, v_i)$. The reward obtained from choosing the upper part of hexagon $H_i$ is exactly $\alpha_i$.
\end{lem}

\begin{proof}
    We start by proving via induction over $i$ that $\pi$ can reach hexagon~$H_i$ (i.e., reach~$a_i$) only between $[y_{i-1}, y_i]$ (we set $y_0$ and $y_1$ to be $0$). 

    \vspace{2mm}
    \noindent
    \textbf{Base ($i=1$):}
    Since $\pi$ starts at the beginning of the first hexagon $H_1$ it reaches~$a_1$ at time $0$ and indeed $0 \in [0, 0] = [y_0, y_1]$.

    \vspace{2mm}
    \noindent
    \textbf{Step:}
    We assume that the induction hypothesis holds for hexagon $H_i$ and we will prove it for hexagon $H_{i + 1}$. 
    $\pi$ can either go through the lower or upper part of hexagon~$H_i$, and following Obs.~\ref{obs:no-wait} we know that $\pi$ can't wait at any vertex of $H_i$.
    If $\pi$ goes through the lower part of hexagon $H_i$, it will reach hexagon $H_{i+1}$ between $[y_{i-1} + \kappa_{i-1}, y_i + \kappa_{i-1}] = [y_i, y_{i+1} - x_i]$.
    If $\pi$ goes through the upper part of~$H_i$, it will reach~$H_{i+1}$ between $[y_{i-1} + \kappa_{i}, y_i + \kappa_{i}] = [y_i~+~x_i, y_{i+1}]$.
    Taking a union over both options, we get that $\pi$ will reach~$H_{i+1}$ between $[y_i, y_{i+1}~-~x_i]~\cup~[y_i~+~x_i, y_{i+1}] = [y_i, y_{i+1}]$ which completes our induction.

    Now we can complete the proof of Lemma~\ref{lma:reward}.
    If $\pi$ traverses the upper part of hexagon $H_i$, it will reach $b_i$ between $[y_{i-1}, y_i]$. 
    Thus, it will reach $c_i$ between $[y_{i-1} +\kappa_i, y_i + \kappa_i] = [y_i + x_i, y_{i+1}]$ which means $\pi$ must be on the edge $(b_i, c_i)$ between $[y_i, y_i + x_i]$ (since $\pi$ never waits at a vertex). In addition $\alpha_i = x_i/\kappa_i \leq 1 \leq x_i$ thus $[y_i, y_i + \alpha_i] \subseteq [y_i, y_i + x_i]$ and since both $b_i$ and $c_i$ have an interval between $[y_i, y_i + \alpha_i]$ we get that $\pi$ must obtain a reward of $\alpha_i$. Since this is the only interval in $H_i$, $\pi$ obtains a reward of exactly~$\alpha_i$.
\end{proof}

\begin{lem}
\label{lma:reach-end}
    The maximal reward obtainable for any path $\pi$ before reaching vertex~$v$ is bounded by $1$.
\end{lem}
\begin{proof}
    The maximal reward that can be obtained without reaching vertex $v$ can be upper bounded by the union of all the intervals that do not belong to $v$. Let~$R_{\max}^{ \not v}$ be the maximal reward obtainable without reaching $v$, it holds that: 
    \begin{align*}
    R_{\max}^{\not v} \leq \sum_{i=1}^{n}{\alpha_i} = \sum_{i=1}^{n}{x_i / \kappa_n} = \frac{1}{\kappa_n} \cdot \sum_{i=1}^{n}{x_i} = 1.
    \end{align*}
\end{proof}

\begin{lem}
\label{lma:max-reward}
    The maximal reward obtainable is exactly $1+ \k / \kappa_n $.
\end{lem}

\begin{proof}
    Let $\pi$ be an optimal path. From Lemma~\ref{lma:reach-end} we know that $\pi$ must reach $v$, and since there is a valid interval on vertex $u$, $\pi$ must reach it before $y_n + K$. 
    
    Let $X'$ be the set of items represented by the hexagons at which $\pi$ chose the upper part, formally: $X' = \{ x_i~|~b_i \in \pi \}$. Following Obs.~\ref{obs:down-path}, \ref{obs:choosing-up} we get that: $\sum_{x_i \in X'}{x_i} \leq \k$. From Lemma~\ref{lma:reward} we get that the reward obtained from choosing the upper part of hexagon $H_i$ is $x_i / \kappa_n$ therefore the maximum reward that can be obtained from the hexagons is $\k / \kappa_n$. We can add the reward we obtain at vertex $v$ and get that the maximal reward is $1+\k / \kappa_n$.
\end{proof}

\noindent We are finally ready to prove our theorem.
\begin{proof}
Let $X' \subseteq X$ be a subset such that $\sum_{x_i \in X'}{x_i} = K$. We build our path~$\pi$ as follows: 
for each $i$, if $x_i \in X'$ append $\langle a_i, b_i, c_i \rangle$ to the path else append $\langle a_i, d_i, e_i \rangle$. Finally, append $u, v$ to the end of the path. 
Next, we build or timing profile $\T = \langle t_0, \dots, t_{k-1} \rangle$ by setting $t_0 := 0$, and $t_i := t_{i-1} + \ell^+(v_{i-1}, v_i)$. We  start by showing that we arrive at vertex $v$ at time $y_n + \k$:
\begin{align*}
t_{k-1} &= t_0 + \sum_{i=1}^{k-1}{\ell^+(v_{i-1}, v_i)} \\
&= \sum_{x_i \in X'}{\kappa_i} + \sum_{x_i \notin X'}{\kappa_{i-1}} \\
&= \sum_{i = 1}^{n}{\kappa_{i-1}} + \sum_{x_i \in X'}{x_i} \\
&= \sum_{i = 1}^{n-1}{\kappa_i} + \sum_{x_i \in X'}{x_i} = y_n + \k
\end{align*}
Thus, our path and timing profile obtain a reward of 1 in the time interval $[y_n+\k, y_n + \k + 1]$. 
We can now analyze the reward obtained during the time interval $[0, y_n + K]$.
From Lemma~\ref{lma:reward}, we get that the reward obtained during the time $[0, y_n + \k]$ is equal to: 
\begin{align*}
\sum_{x_i \in X'}{\alpha_i} &= \sum_{x_i \in X'}{{x_i} / {\kappa_i}} \\
&= \frac{1}{\kappa_n} \cdot \sum_{x_i \in X'}{x_i} \\
&= {\k} / \kappa_n 
\end{align*}
Thus, the total reward of our path and timing profile is $1+\k / \kappa_n$ and from Lemma~\ref{lma:max-reward} we know it is the maximal reward thus our path and timing profile are optimal. 

Now, let $\pi, \T$ be an optimal path and timing profile such that $\R(\pi, \T) = 1+\k / \kappa_n$. From Lemma~\ref{lma:reach-end} we know that $\pi$ must reach vertex $v$, and since it is possible to reach $v$ only before $y_n + \k$ we know that $\pi$ earns a reward of~$1$ at $v$. Thus, it must earn a reward of $\k / \kappa_n$ until $v$. Let $X'$ be the set of items represented by the hexagons at which $\pi$ chose the upper part. Formally, $X' = \{ x_i~|~b_i \in \pi \}$. From Lemma~\ref{lma:reward} we know that the reward obtained from choosing the upper part of hexagon $i$ is exactly ${x_i} / \kappa_n$ thus:
\begin{align*}
    \k / \kappa_n &= \sum_{x_i \in X'}{x_i / \kappa_n} \\
    \Rightarrow \k &= \sum_{x_i \in X'}{x_i},
\end{align*}
which completes our reduction.
\end{proof}

\section{OPTP Upper Bounds---Additional Details}
\label{app:bound}

We start by formally defining $\delta'(u,v)$.
We then describe in detail our algorithm used to compute $\ub_I$ which was briefly described in Sec~\ref{subsec:optp-bounds}. 
We then proceed to prove Lemmas~\ref{lem:ub-i},~\ref{lem:ub-ut} and Thm.~\ref{thm:ub-pi}.

\begin{dfn}
    Let $G = (V, E)$ be a task-assistance graph. We define $\delta'(u,v)$ to be the minimum distance between $u$ and $v$ in $G$ while not accounting for half of the first and last edge. 
    Namely, if $\Pi(u,v)$ is the set of all paths from $u$ to $v$ then 
    \[
        \delta'(u,v) := \min_{\pi \in \Pi(u,v)}\{\ell_\pi^-(v) - \ell_\pi^+(u)\}
    \]
    if $\Pi(u,v) \neq \emptyset$ and $\infty$ if no such path exists.
\end{dfn}

\begin{algorithm}
\caption{compute intervals upper bounds}\label{alg:ub_i}
\hspace*{\algorithmicindent} 
    \textbf{Input:} graph $G = (V,E)$;
    \hspace{3mm} 
    intervals:  $\mathcal{I}$ \\ 
    \hspace*{5mm}
    \textbf{Output:} 
    upper bound $\ub_I$ for each interval $I$
    
\begin{algorithmic}[1]

\State $\intervalset \gets \emptyset$  \label{alg2:init-start} \cmm{gather all intervals into a set}
\For {$u \in V$} 
    \For {$I=[t_s,t_e] \in \mathcal{I}(u)$}
        \State \intervalset.insert($I$);  
        \State $\ub_I \gets \vert I \vert$ \cmm {initialize bound for $I$}
    \EndFor 
\EndFor
\State $N_\calI^G \leftarrow \vert \intervalset \vert $ \label{alg2:init-end}
\vspace{2mm}
\Loop{~$N_\calI^G$ times} \label{alg2:outer-start}
    \For {$I=[t_s,t_e] \in \intervalset$}
        \State $u \gets I$.vertex() \cmm {vertex that $I$ belongs to}
        \For {$I'=[t_s',t_e'] \in \intervalset$}
            \State $v \gets I'$.vertex() \label{alg2:reward-start} \cmm {vertex that $I'$ belongs to}
            \If{$I'$ is reachable from $I$}
                \State $t_{\text{unused}} \gets \max \{ 0, t_e + \delta'(u,v) - t'_s \}$ 
                \cmm {time of $I, I'$ that cant be used}
                \State $\ub_I \gets \max\{\ub_I, \vert I \vert + \ub_{I'} - t_{\text{unused}} \}$\label{alg2:update-reward}
            \EndIf
        \EndFor \label{alg2:reward-end}
    \EndFor
\EndLoop    \label{alg2:outer-end}
\State \Return {$\{ \ub_I~\vert~I \in \intervalset \}$}

\end{algorithmic}
\end{algorithm}

Alg.~\ref{alg:ub_i} provides the pseudo code to compute $\ub_I$. 
It starts by initializing $\intervalset$ to be a set containing all intervals,
$\ub_I$ to be the length of interval~$I$ 
and $N_\calI^G$ to be the number of intervals in $G$ 
(Lines~\ref{alg2:init-start}-\ref{alg2:init-end}).
It then iterates over all intervals $N_\calI^G$ times (Lines~\ref{alg2:outer-start}-\ref{alg2:outer-end}). 
For each interval $I$, the algorithm iterates over all intervals $I'$ and bounds the maximal reward obtainable assuming the next interval after $I$ is $I'$ (Lines~\ref{alg2:reward-start}-\ref{alg2:reward-end}). 
This is done by computing $t_{\text{unused}}$, which is the portion of either $I$ or $I'$ in which no reward can be obtained when traveling from $I$ to $I'$, and subtracting it from $\vert I \vert + \ub_{I'}$.
If the bound obtained is greater then the current $\ub_I$, the algorithm updates $\ub_I$ accordingly (Line~\ref{alg2:update-reward}).

\begin{replemma}{lem:ub-i}
    Let $\pi$, $\T$ be a path and timing-profile. Let $I=[t_s,t_e]$ be an interval~$\T$ obtains reward from. Then it holds that $\ub_I \geq \R(\pi, \T, t_s)$.
\end{replemma}

\begin{proof}
    Let $I_0, \dots, I_m$ be the list of intervals $\T$ obtains reward from ordered from last to first. We will prove the lemma by induction over $i$ such that at the end of every iteration $i$, it holds that $\ub^i_{I_i} \geq \R(\pi, \T, t_s)$ when $I_i = [t_s, t_e]$. 

    \vspace{2mm}
    \noindent
    \textbf{Base ($i=0$):}
    Since $I_0 = [t_s, t_e]$ is the last interval $\T$ obtains reward from it holds that $\R(\pi, \T, t_s) \leq \vert I_0 \vert = \ub^0_{I_0}$.

    \vspace{2mm}
    \noindent
    \textbf{Step:}
    We assume that the induction hypothesis holds for $i-1$ and we will prove for $i$.
    Let $u$ be the vertex $I_{i} = [t_s, t_e]$ belongs to, and $v$ the vertex $I_{i-1} = [t_s', t_e']$ belongs to. 
    $\T$ can not earn reward for at least $t_{\text{unused}} := \max \{0, t_e + \delta'(u,v) - t'_s \}$ time when moving from $u$ to $v$. Thus,
    \[
        \R(\pi, \T, t_s) \leq \vert I \vert + \R(\pi, \T, t'_s) - w.
    \]
    From the induction assumption we get that
    \[
        \R(\pi, \T, t_s) \leq \vert I \vert + \ub^{i-1}_{I_{i-1}} - w.
    \]
    And after the end of the $i$-th iteration it holds that
    \[
        \ub^i_{I_i} \geq \vert I \vert + \ub^{i-1}_{I_{i-1}} - w.
    \]
    Thus, $\ub^i_{I_i} \geq \R(\pi, \T, t_s)$ which concludes the proof.

\end{proof}

\begin{replemma}{lem:ub-ut}
    Let $u \in V$ be a vertex, and $t \in [0,1]$. 
    For every path $\pi$ and timing-profile $\T$
    s.t. $u \in \pi$ and that following $\T$ implies that at time $t$ the robot is at vertex $u$, it holds that $\ub(u, t) \geq \R(\pi, \T, t)$. 
\end{replemma}

\begin{proof}

Let $I = [t_s, t_e] \in \calI(v)$ for some $v \in V$ be the first interval $\T$ obtains reward from after time $t$, and let $t' \geq t$ be the time $\T$ enters that interval (or $t$ if $t' < t$).
Since $\delta'(u,v)$ is the minimal time with no reward between $u$ and $v$ it holds that $t' \geq t + \delta'(u,v)$.
Recall that $t_{\text{unused}} := \max \{ 0, t + \delta'(u, v) - t_s \}$ is the time that must pass since $t_s$ in which $\T$ can not obtain reward from $I$, 
then from Lemma~\ref{lem:ub-i} it holds that $\R(\pi, \T, t) \leq \ub_I - t_{\text{unused}}$, 
And from its definition, it holds that $\ub(u,t) \geq \ub_I - t_{\text{unused}}$,
thus, we get that $\ub(u,t) \geq \ub_I - t_{\text{unused}} \geq \R(\pi, \T, t)$.

\end{proof}

\begin{reptheorem}{thm:ub-pi}
    Let $\pi = \path$ be a path. For any path $\pi' = \langle v_0, \dots, v_k, \dots, v_m \rangle$ extending $\pi$ s.t., $\T'_{\pi'} = \langle t'_0, \dots, t'_{m-1} \rangle$ is its optimal timing profile, it holds that   $\ub(\pi) \geq \R(\pi', \T'_{\pi'})$.
\end{reptheorem}

\begin{proof}
    Let $I = [t_s, t_e] \in \calI(v_i)$ for some $v \in \pi$ be the last interval $\T'$ obtains reward from before leaving $v_k$.
    We separate the reward obtained by $\pi'$'s into two parts: 
    (i)~the reward obtained until $t'_i$ (the time $\T'$ exits $v_i$)
    and 
    (ii)~the reward obtained from~$t'_i$.

    Since $v_i \in \pi$, it holds that $t_e$ is a Vertex-critical time of vertex $v_i$ in $\pi$.
    Thus, given the time-reward pair $(t_e, r)$ belonging to the \exit list of $v_i$ when running the OTP algorithm on $\pi$\footnote{
    In order for the next part to be correct in the case where $t'_i > t_e$, we must consider entering vertex $v_i$ at time $t'_{i-1} > t_e - \ell^+(v_{i-1}, v_{i})$ which is not considered when computing the pair $(t_e, r)$. To do so, we update the ``if'' statement (Line~\ref{alg-cbr:if}) in \texttt{compute\_best\_reward} (Alg.~\ref{alg:cbr}) to be: ``if $t_{\text{entry}} > \texit$'' when running the OTP algorithm for $\ub(\pi)$. 
    }
    it holds that $\R(\pi', \T') - \R(\pi', \T', t_e) \leq r$, and since $\R(\pi', \T', t_e) \leq \R(\pi', \T', t'_i)$ we get that $\R(\pi', \T') - \R(\pi', \T', t'_i) \leq r$.

    Additionally, $\T'$ does not obtain reward from $t'_i$ to $t'_k$, and it holds that $t'_k \geq t'_i + \ell^+(v_i, v_k) \geq t_s + \ell^+(v_i, v_k)$. Thus, from Lemma~\ref{lem:ub-ut} we get that $\R(\pi', \T', t'_i) \leq \ub \left(v_k, t_s + \ell^+(v_i, v_k)\right)$.
    Finally:
    \begin{align*}
    \R(\pi', \T') &= \R(\pi', \T') - \R(\pi', \T', t'_i) + \R(\pi', \T', t'_i) \\
                  &\leq r + \ub \left(v_k, t_s + \ell^+(v_i, v_k) \right) \leq \ub(\pi).
    \end{align*}
\end{proof}

\end{document}